\documentclass[empt,nonblindrev]{informs3_mb_ver02}

\usepackage[round]{natbib}
\usepackage{booktabs}
 \bibpunct[, ]{(}{)}{,}{a}{}{,}%
 \def\bibfont{\small}%
\TheoremsNumberedThrough     
\ECRepeatTheorems

\EquationsNumberedThrough    

\def\sl{\em}
\def\reals{\mathbb{R}}

\def\E{\mathbb{E}}
\def\P{\mathbb{P}}
\def\cF{\mathcal{F}}
\def\bfX{\mathbf{X}}
\def\myep{\varepsilon}
\def\vx{\mathbf{x}}
\def\vy{\mathbf{y}}
\def\vq{\mathbf{q}}
\def\xmax{x_{\max}}
\def\rmax{r_{\max}}
\def\trans{\top}
\def\lmin{\lambda_{\min}}
\def\lmax{\lambda_{\max}}

\def\naturals{\mathbb{N}}
\usepackage{graphicx}
\usepackage{caption}
\usepackage{subcaption,hyperref}

\begin{document}


\RUNAUTHOR{Qiang and Bayati}

\RUNTITLE{Dynamic Pricing with Demand Covariates}

\TITLE{Dynamic Pricing with Demand Covariates}

%

\ARTICLEAUTHORS{%
\AUTHOR{Sheng Qiang}
\AFF{Stanford University Graduate School of Business, Stanford, CA 94305, \EMAIL{sqiang@stanford.edu}} 
\AUTHOR{Mohsen Bayati}
\AFF{Stanford University Graduate School of Business, Stanford, CA 94305, \EMAIL{bayati@stanford.edu}} %
%
} 

\ABSTRACT{%
We consider a generic problem in which a firm sells products over $T$ periods without knowing the demand function. The firm sequentially sets prices to earn revenue and to learn the underlying demand function simultaneously. A natural heuristic for this problem, commonly used in practice, is greedy iterative least squares (GILS). At each time period, GILS estimates the demand as a linear function of the price by applying least squares to the set of prior prices and realized demands. Then a price that maximizes the revenue, given the estimated demand function, is used for the next time period. The performance is measured by the regret, which is the expected revenue loss from the optimal (oracle) pricing policy when the demand function is known. Recently, \cite{den2014simultaneously} and \cite{keskin2014dynamic} demonstrated that GILS is sub-optimal. They introduced algorithms which integrate forced price dispersion with GILS and achieve asymptotically optimal performance.

In this paper, we consider this dynamic pricing problem in a data-rich environment. In particular, we assume that the firm knows the expected demand under a particular price from historical data, and in each period, before setting the price, the firm has access to extra information (demand covariates) which may be predictive of the demand. Demand covariates can include marketing expenditures, geographical information, consumers socio-economic attributes, macroeconomic indices, weather, etc. We prove that in this setting the behavior of GILS is dramatically different and it achieves asymptotically optimal regret of order $\log(T)$. We also show the following surprising result: in the original dynamic pricing problem of \cite{den2014simultaneously,keskin2014dynamic}, inclusion of any set of covariates in GILS as potential demand covariates (even though they could carry no information) would make GILS asymptotically optimal. We validate our results via extensive numerical simulations on synthetic and real data sets.

}%


\KEYWORDS{dynamic pricing, exploration and exploitation, demand covariates}
\HISTORY{This version appeared on April 14, 2016.}

\maketitle

%

%

\section{Introduction}\label{sec:intro}

Companies launch new products periodically without accurate prior knowledge of the true demand. One method to learn the demand function is price experimentation, where companies adaptively modify prices to learn the hidden demand function and use that to estimate the revenue-maximizing price. Nevertheless, proper price experimentation is a challenging problem due to the potential revenue loss during the learning horizon that can be substantially large. In fact optimally balancing the trade-off between randomly selecting prices to expedite the \emph{learning} versus selecting prices that maximize the expected \emph{earning} has been subject of recent research in the operations management literature. Learning and earning are also known by exploration and exploitation respectively.
In particular, \cite{den2014simultaneously,keskin2014dynamic} studied greedy iterated least squared (GILS), a popular heuristic used in practice, which works by greedily selecting the best price based on the most up-to-date estimation of the demand at any time. GILS pricing policy is also known by \emph{myopic pricing} or \emph{certainty equivalent pricing}. \cite{den2014simultaneously,keskin2014dynamic} showed that GILS is sub-optimal and suffers from \emph{incomplete learning}. In addition, they introduced algorithms that integrate forced price dispersion with GILS and achieve asymptotically optimal regret. Despite these results, in most applications, greedy policies such as GILS are popular due to their simplicity and due to the common perception of the firms that price experimentation could lead to substantial revenue loss.

On the other hand, thanks to growing availability of data and advances in statistical learning, estimation problems such as the aforementioned demand learning can be solved much more accurately in data-rich environments. The most common approach in boosting the accuracy is through introduction of covariates (or predictors or features) and finding a function that accurately maps the covariates and price to the outcome of interest (demand in our case). Examples of covariates for the demand estimation could be market size, macroeconomic indices, seasonality, or geographic indicators.

In this paper we study the impact of including demand covariates on the aforementioned trade-off between learning and earning. \emph{While adding these covariates is a natural step to improve the demand estimation and hence to expedite the learning, we show an additional and surprising benefit; inclusion of the demand covariates in the popular GILS approach automatically solves the incomplete-learning problem}. In other words, no pro-active learning is required and the companies can focus all of their attention to earning.
We provide general sufficient conditions for asymptotically optimality of GILS with demand covariates. A technical part of our analysis is proving sharp concentration inequalities for the minimum singular value of the least squared design matrix and relies on a recent concentration inequality for sum of matrix martingales by \cite{tropp2011user}.

In addition, we prove that even if the demand covariates are irrelevant and do not improve the demand estimation, their inclusion is beneficial and makes GILS asymptotically optimal. In other words, the process of learning which covariates have coefficients equal to zero (feature selection) which is usually performed separately by data scientists, if combined with GILS pricing policy, dramatically improves the performance. The added uncertainty from not knowing which variables are predictive, creates automatic price experimentation that solves the incomplete learning problem of GILS.

Finally, via extensive numerical simulations on synthetic and real data sets, we demonstrate robustness of our results to the assumptions required by the theory.

\subsection{Organization of the paper}

The remainder of the paper is organized as follows. Section \ref{sec:lit-rev} surveys related research and it is followed by formal definition of the problem and GILS policy in Section \ref{sec:problem_formulation}. Analysis of the regret and our main theoretical results are presented in Section \ref{sec:regret_analysis} while empirical simulations are given in Section \ref{sec:simulations}. Section \ref{sec:discussion} covers final remarks and discussions, and all proofs are provided in Appendix \ref{sec:proofs}. In Appendix \ref{sec:extensions} we explain how the proofs can be extended when our modeling assumptions on covariates and demand uncertainty are generalized.

%
\section{Related Literature}\label{sec:lit-rev}



The trade-off between learning and earning has long been a focus of attention in statistics \citep{lai1979adaptive,lai1985asymptotically}, computer science \citep{auer2002nonstochastic,auer2003using}, and economics \citep{segal2003optimal,kreps2014}. There has also been recent interest in this topic in the operations management community, especially in the field of dynamic pricing and revenue management. An admittedly incomplete list of such papers is
\citep{kleinberg2003value,carvalho2005learning,Araman2009,besbes2009dynamic,besbes2011minimax,harrison2012bayesian,
den2014simultaneously,keskin2014dynamic,johnson2015online,Chen2015Nonparametric}. We defer the reader to the recent survey by \cite{den2015Survey} and references therein for a complete list and thorough discussion.


Among these, the most related papers with our paper are \citep{den2014simultaneously,keskin2014dynamic}. \cite{keskin2014dynamic} study the balance between learning and earning if the expected demand is a linear function of the price while \cite{den2014simultaneously} consider a more general case-- when the expected demand is a generalized linear function of the price. These papers propose novel variants of the greedy policy, namely controlled variance pricing (CVP) by \cite{den2014simultaneously} and constrained iterative least squared (CILS) by \cite{keskin2014dynamic}, which enforce price dispersion within GILS and achieve asymptotically optimal regret. \cite{keskin2014dynamic} also provide lower bounds for any pricing policy. Our work is closer to \citep{keskin2014dynamic}, since similar to them, we study the situation where demand at a single incumbent price is known to the firm with high accuracy. Our main contributions compared to these papers is studying the problem in a data-rich environment (when demand covariates are available), showing how this addition fundamentally changes behavior of GILS, and that it does not need any forced price dispersion like in CVP and CILS. Similar to \citep{keskin2014dynamic}, we also show a lower bound for performance of any pricing policy in presence of demand covariates. Hence we show that GILS is asymptotically optimal.

This result is surprising since as shown by \cite{lai1982iterated}, the parameter estimates computed under an iterative least squared policy may not be consistent, and the resulting controls are sub-optimal. This fact is also shown by \cite{den2014simultaneously,keskin2014dynamic} in the dynamic pricing problem (without demand covariates) and is termed \emph{incomplete learning} which motivated the novel policies CVP and CILS. Although, it is worth noting that a certain Bayesian form of iterative least squared (when parameters to estimate have a probability distribution with a prior) converges to the optimum values \citep{Chen1998}.

Another related setting is dynamic pricing with demand uncertainty where the demand function is not static and changes over time. Variants of this setting have been studied by \cite{den2015changingenvironment,keskin2013chasing}. Although, at a first look one can consider our demand function that depends on price and covariates as a changing demand function, however our setting is a substantially different problem. The reason is that the information provided by the demand covariates in our model is hidden in demand shocks. In particular, if the covariate information is not available to the firm, our setting can be exactly mapped to the case of static dynamic pricing where demand shocks have higher variance.

In the learning and earning framework, beyond dynamic pricing, there are few papers that consider how covariates may help the decision maker to improve the performance. This problem was first studied in the statistics literature by \cite{woodroofe1979one,sarkar1991one} where a simple one-armed bandits problem with a covariate was considered and they showed that a myopic policy is asymptotically optimal. Much later, a machine learning paper by \cite{langford2007epoch} considered a more general problem of learning and earning with covariates. Their problem was inspired by online advertising where a platform such as Facebook, Google, MSN, or Yahoo needs to match publishers to viewers in order to maximize the expected number of user clicks. They modeled the problem as a $K$-armed bandits problem, with the contexts (covariates) which may predict the revenue of each arm. This pioneered an active research area (contextual bandits). Examples of papers studying contextual bandits are \citep{dudik2011efficient,chu2011contextual,seldin2011pac,li2014contextual,badanidiyuru2014resourceful,
goldenshluger2007performance,goldenshluger2011note,goldenshluger2013linear,rigollet2010nonparametric,perchet2013multi,
lassoBandit2015}. The main difference between these papers and our paper is that in our model the reward is a quadratic function of the action (price) and covaraites. But these papers either consider $K$-armed bandits or linear bandits where the reward function is linear in covariates and the action. In addition, none of these contextual bandit papers show optimality results for the greedy policy which is popular among practitioners. In fact, the greedy-type policies studied by these paper, similar to CVP and CILS, have a degree of forced experimentation where greedy policy is used in $1-\epsilon$ fraction of the periods to maximize the reward and the remaining $\epsilon$ fraction of periods are dedicated to forced experimentation to improve learning. The value of $\epsilon$ is reduced over time to avoid a linear regret.

Our results are also related to the recent growing literature in Operations Management that studies benefits of combining statistical estimation with decision optimization \citep{Shanthikumar2005,Levi2015,Rudin2014Vahn,kallus2015Bertsimas}. The main difference is that these papers focus on a static decision making task while here we are considering a dynamic decision making problem. In particular, our focus is on the impact of combining statistical estimation with price optimization on creating enough price dispersion (exploration) which is required for the pricing policy.

While writing this paper, we became aware of a very recent paper by \cite{Cohen2016featurebased} that also studies dynamic pricing with covariates. For the following reasons the two papers are fundamentally different. \cite{Cohen2016featurebased} assume that demand is equal to $1$ if price is less than a linear function of the covariates which is interpreted as valuation of the customer. If the price is greater than the valuation then demand is $0$. But our demand function is quadratic in price and linear in covariates. In addition, our demand function includes stochastic demand shocks. Due to the threshold form of the demand function in \citep{Cohen2016featurebased}, estimating the problem parameters (coefficients of the covariates) can no longer be solved by a simple procedure such as least squared. On the other hand, their demand function is deterministic and does not contain demand shocks which are the core reason behind the difficulty of demand estimation studied by us. \cite{Cohen2016featurebased} consider adversarial covariates while our covariates are stochastic. Finally, we provide both upper bounds and lower bounds for the regret of a popular algorithm in practice (GILS) while \cite{Cohen2016featurebased} introduce new algorithms for their setting and prove only upper bounds for the regret. Another paper at the intersection of dynamic pricing and contextual bandits is by \cite{Amin2014ContextualAuction} which considers a problem similar to \cite{Cohen2016featurebased} that is different from ours.

%

\section{Problem Formulation}\label{sec:problem_formulation}

Consider a firm that sells one type of product over time. In each period, the firm can adjust price of the product. Customer demand for the product is determined by the price and some other factors (demand covariates), according to an underlying parametric demand function. The firm is initially uncertain about the parameters of the demand function but can use historical data on charged prices, realized demands, and observed values of the demand covariates to estimate the parameters. This is the problem of \emph{dynamic pricing with demand covariates}.

\subsection{Model Setting}

Our model builds on the model of \cite{keskin2014dynamic}.
We assume the firm sells the product over a time horizon of $T$ periods. The total periods $T$ is a large number and is unknown to the seller, which prevents the firm from using $T$ as a decision factor. In each period $t \in \naturals$, where $\naturals$ is the set of positive integers, the seller observes some demand covariates of the market, that is a vector of covariates $\vx_t=(x_{1t}, x_{2t},\dots, x_{mt})\in\reals^{m}$ sampled independently from a fixed distribution $p_{\vx}$. We assume that $p_{\vx}$ is absolutely continuous with respect to the Lebesgue measure on $\reals^m$ and has compact support, i.e. there is $\xmax\in\reals^+$ such that $\|\vx_t\|_\infty\le \xmax$ for all $t$ in $\naturals$.
Then the seller must choose the price $p_t$ from a given feasible set $[l,u] \subset \reals$, where $0 \le l < u < \infty$. After that, the seller observes the demand $D_{t}$ in period $t$. We assume the demand follows a linear function of the prices and covariates, which is commonly used in economic literature to illustrate the relationship between demand and price.
Hence, the demand function is formulated as
\begin{eqnarray}\label{eq:demand}
D_{t} = \alpha + \beta p_t + \gamma \cdot \vx_t + \myep_{t} \qquad\hbox{for all } t \in \naturals
\end{eqnarray}
where $u\cdot v$ denotes the inner product of vectors $u$ and $v$, $\alpha \in \reals$ is the basic market size, $\beta \in \reals$ is the price coefficient (that is $\beta <0$), and $\gamma = (\gamma_{1},\dots,\gamma_{m})^{\trans}\in \reals^{m}$ is a vector of coefficients for the demand covariates. The parameters $\alpha, \beta, \gamma$ are assumed to be unknown to the seller. In addition, $\{\myep_{t}\}_{t\ge 1}$ are unpredictable demand shocks, drawn independently identically from a distribution with zero mean, finite variance denoted by $\sigma_\myep^2$. Note that we assume there is a large enough inventory that any amount of demand can be fulfilled by the firm.

Next, without loss of generality, we assume that $\E_{p_{\vx}}[\vx]=0$ since any possible non-zero expectation can be moved to the intercept term $\alpha$. Additionally, we assume that the covariance matrix of demand covariates (denoted by $\Sigma_\vx$) is positive definite. To make the math cleaner, we also assume that each coordinate of $\gamma$ is re-scaled so that the diagonal entries of $\Sigma_\vx$ are all equal to $1$ (i.e., all coordinates of $\vx$ have unit variance).
\begin{remark}\label{rem:randomness-of-x-ep}
In Section \ref{sec:non-iid} we will show that our main result (upper bound on performance of GILS with demand covariate) holds when the i.i.d. assumptions on sequences $\{\myep\}_{i\ge1}$ and $\{\vx_i\}_{i\ge1}$ are replaced by weaker assumptions that they are martingale differences with respect to the past.
\end{remark}
We also assume that the seller has access to one more piece of prior information; through extensive use of an incumbent price $p_0$ in the past the seller knows the average demand at this incumbent price with high accuracy. More precisely, denoting the average historical demand as $\bar{D}$, the fact that shocks and $\vx$ have zero average help us to simplify the demand function because the expected demand satisfies $\E[D] = \alpha + \beta p_0$. Therefore, with large number of uses of the incumbent price $p_0$, the average value is close to its expected value by the strong law of large numbers. Thus, we can write
$\bar{D} = \alpha + \beta p_0$ which leads to the following modified demand function
\begin{eqnarray*}
D_{t} = \bar{D} + \beta (p_t - p_0) + \gamma \cdot \vx_t + \myep_{t} \qquad\hbox{for all } t \in \naturals\,.
\end{eqnarray*}
We can treat $\bar{D}$ as a given number $a'$. Hence, are final demand model is
\begin{eqnarray}\label{eq:DP-incumbent}
D_{t} = a' + \beta (p_t - p_0) + \gamma \cdot \vx_t + \myep_{t}\qquad\hbox{for all } t \in \naturals\,,
\end{eqnarray}
in which $a'$ is known, and $(\beta, \gamma)$ are unknown parameters.
\begin{remark}\label{rem:incumbent}
Throughout the rest of the paper when we use the phrase ``dynamic pricing problem'' it is implicitly assumed that we are referring to the case where an incumbent price $p_0$ is available, or equivalently, to problem \eqref{eq:DP-incumbent} with known $a'$.
\end{remark}
To simplify the notation, we let
\[
\theta\equiv
\begin{pmatrix}
\beta\\
\gamma
\end{pmatrix}
\in \reals^{m+1}\,
\]
be the column vector of demand model parameters, and express the demand vector $\mathbf{D}_t = (D_{1}, D_2,\dots,D_{t})^{\trans}$ in $\reals^{t}$ in terms of $\theta$ as follows
\begin{eqnarray}\label{eq:dem}
\mathbf{D}_t = a'\cdot \mathbf{1}_t + Z_t\theta + \mathbf{e}_t\,.
\end{eqnarray}
Here $\mathbf{e}_t \equiv (\myep_1, \myep_2, \dots, \myep_t)^{\trans} \in \reals^{t}$, $\mathbf{1}_t \equiv (1,\dots, 1)^{\trans} \in \reals^{t}$, and $Z_t$ is a $t \times (m+1)$ matrix with its $i^{\rm th}$ row equal to $u_i^{\trans}$ where
\[
u_i = {p_i -p_0 \choose \vx_i}\in\reals^{(m+1)}\,.
\]
Without loss of generality, we assume that the prices include subtraction of production cost and we do not write them explicitly in the formula. Thus, the terms profit and revenue are used interchangeably. The seller's expected single-period revenue function is
\begin{eqnarray}
r_{\theta}(p,\vx)\equiv p\,[a' + \beta (p - p_0) + \gamma\cdot \vx]\,.
\end{eqnarray}
Next we define $p^*(\theta,\vx)$ to be the price that maximizes the expected single-period revenue function $r_{\theta}(\cdot,\vx)$ given the demand covariates $\vx$, i.e.
\begin{eqnarray}
p^*(\theta,\vx) \equiv\arg\max_{p\in [l,u]}\left[r_{\theta} (p,\vx)\right]\,.
\end{eqnarray}
We assume that the true parameter $\theta$ is in the compact set
\[
\Theta \equiv \Big\{
\begin{pmatrix}
\beta\\
\gamma
\end{pmatrix}
\in \reals^{m+1}
~\Big|~ -\infty < b_{\min} \le \beta \le b_{\max} <0\,,
\textrm{ and } \| \gamma \| \le \rmax <\infty\,
\Big\}\,,
\]
where $\|.\|$ is the $\ell_2$ norm. The condition on $\beta$ just means that the expected demand is strictly decreasing in price and we have an uncertain interval of negative numbers around it. We also assume that the optimal price corresponding to any such true parameter $\theta$ and any vector of demand covariates $\vx$ is an interior point of the feasible set $[l,u]$. Similar assumption was made by \cite{den2014simultaneously} and \cite{keskin2014dynamic}. The reason for the assumption is to avoid having the optimal price to be a corner solution of the interval $[l,u]$ so we can use first order conditions. In particular, the first-order condition for optimality would be
\begin{eqnarray}
a' + 2\beta p - \beta p_0 + \gamma \cdot \vx = 0\,,
\end{eqnarray}
from which we deduce that
\begin{eqnarray}
p^*(\theta,\vx) = \frac{a' + \gamma\cdot \vx}{-2\beta} + \frac{p_0}{2}\,.
\end{eqnarray}
For the incumbent price $p_0$, we consider there is a small positive constant $\delta_0 > 0$, such that
\begin{equation}\label{eq:delta-condition}
\frac{a'}{-b_{\max}} - p_0 \ge \delta_0~~~\textrm{or}~~~p_0 - \frac{a'}{-b_{\min}} \ge \delta_0\,.
\end{equation}
This condition will guarantee that for any $\tilde{\beta}\in (b_{\min},b_{\max})$ the following inequality holds,
\[
\left|\frac{a'}{-2\tilde{\beta}}-\frac{p_0}{2}\right|\ge\delta_0\,.
\]
%

%
%
\subsection{Pricing Policies and Performance Metric}

Let $H_{t-1}$ denote the observed demands, prices and the demand covariates before choosing a price and realizing demand at period $t$. That is, $H_{t-1}\equiv (D_1,\dots,D_{t-1},p_1,\dots,p_{t-1},\vx_1,\dots,\vx_{t})$. We define a pricing policy as a sequence of functions $\pi = (\pi_1,\pi_2,\dots)$, where %
\[
\pi_t: H_{t-1} \rightarrow [l,u]
\]
for all $t=2,3,\dots$, and $\pi_1$ is a deterministic function of $\vx_1$.
Now, we clarify the probability space for the performance. Any pricing policy induces a family of probability measures on the sample space of demand sequences $(D_1, D_2,\dots)$ as below. Given the parameter vector $\theta$, the realized demand covariates $X_{1:T}\equiv\{\vx_1,\vx_2,\dots,\vx_T\}$, and a pricing policy $\pi$, let $\mathbb{P}_{\myep}(\cdot|X_{1:T})$ be the probability measure with respect to the randomness of demand shocks, i.e.
\begin{eqnarray}
\mathbb{P}_{\myep} (D_1\in d \xi_1,\dots,D_T \in d \xi_T|X_{1:T})
= \prod_{t=1}^{T}\mathbb{P}_{\myep_t} (a' +\beta p_t + \gamma\cdot \vx_t + \myep_{t} \in d \xi_{t}|X_{1:t} )\,.
\end{eqnarray}
Thus, $p_t = \pi_t(H_{t-1})$, which implies that $p_t$ is completely characterized by $\pi$, $\theta$, and $H_{t-1}$.
Accordingly the  $T$-period expected revenue of the seller is
\begin{eqnarray}
R_\theta^\pi (T; X_{1:T}) \equiv \E_{\myep} \left[ \left. \sum_{t=1}^{T} r_\theta(p_t,\vx_t) \right| X_{1:T} \right]\,,
\end{eqnarray}
where $\E_{\myep}$ is the expectation operator with respect to the randomness in demand shocks. The performance metric we will use throughout this study is the $T$-period regret, which is a random variable depending on the realized demand covariates sequence $X_{1:T}$, defined as
\begin{eqnarray}
\Delta_\theta^\pi (T; X_{1:T}) \equiv \sum_{t=1}^T r_\theta^*(\vx_t) - R^\pi_\theta (T; X_{1:T}) \quad \hbox{ for } \theta \in \Theta \hbox{ and } T \in\naturals\,,
\end{eqnarray}
where $r^*_\theta(\vx_t)\equiv r_\theta\big(p^*(\theta,\vx_t),\vx_t\big)$ is the optimal expected single-period revenue, given the demand covariates $\vx_t$. After algebraic manipulation of the above expression for the regret, we can write the $T$-period regret as
\begin{eqnarray}
\Delta_\theta^\pi (T; X_{1:T}) = -\beta \sum_{t=1}^{T} \E_{\myep}  \left\{ [p_t - p^*(\theta,\vx_t)]^2\left.\right| X_{1:t}\right\}\,.
\end{eqnarray}
While deriving our lower bounds on best achievable performance, we will also make use of the worst-case regret, which is
\begin{eqnarray}
\Delta^\pi (T; X_{1:T}) = \sup_{ \theta\in\Theta}\Delta^\pi_\theta(T; X_{1:T})\quad \hbox{ for } T\in\naturals\,.
\end{eqnarray}
%
The regret of a policy can be interpreted as the expected revenue loss relative to a clairvoyant policy that knows the true value of $\theta$ at the beginning; smaller values of regret are more desirable for the seller.

\subsection{Greedy Iterated Least Squares (GILS)}

Given the history of demands, prices, and demand covariates through the end of period $t$, the least squares estimator of $\theta$ is given by
\begin{eqnarray}\label{eq:LS}
\widehat{\theta}_t \equiv \arg\min_\theta \left[\sum_{s=1}^{t} \| D_s - (a' + \beta p_s + \gamma \cdot \vx_s)  \|_2^2\right]\,,
\end{eqnarray}
for $\theta=(\beta, \gamma^{\trans})^{\trans}$. Using the first-order optimality conditions of the least squares problem (\ref{eq:LS}),  the estimator $\widehat{\theta}_t$ can be expressed explicitly. In particular, the simplified demand function (\ref{eq:dem}) gives us
\begin{eqnarray}
\widehat{\theta}_t = (Z_t^{\trans} Z_t)^{-1} Z_t^{\trans} (\mathbf{D}_t - a'\cdot \mathbf{1}_t)\,,
\end{eqnarray}
and the deviation from the true parameter is
\begin{eqnarray}
\label{eq:dev}
\widehat{\theta}_t - \theta = (Z_t^{\trans} Z_t)^{-1} Z_t^{\trans} \mathbf{e}_t\,.
\end{eqnarray}
Because $\theta$ lies in the compact set $\Theta$, the accuracy of the unconstrained least squares estimate $\widehat{\theta}_t$ can be improved by projecting it into the set $\Theta$. We denote by $\vartheta$ the truncated estimate that satisfies $\vartheta_t \equiv \arg \min_{\vartheta\in\Theta} \{ \|\vartheta - \widehat{\theta}_t\| \}$, where by assumption the corresponding price vector $p^*(\vartheta_{t-1},\vx_t)$ is an interior point of the feasible set $[l,u]$. We call the policy that charges price $p_t = p^*(\vartheta_{t-1},\vx_t)$ in period $t \in\naturals$ the greedy iterated least squares (GILS) policy with demand covariates and denote it by $\pi^g$. Throughout the rest of the paper we make the following assumption.
\begin{assumption}\label{assump:1}
We only consider the subclass of GILS policies where the initial $m+1$ prices are selected so that the vectors
$[p_s~~ \vx_s^{\trans}]^{\trans}$ for $s\in \{1,2,\ldots,m+1\}$ are linearly independent.
\end{assumption}
Note that this is not a restricting assumption. In fact, if the $m+1$ initial prices are selected from any absolutely continuous distribution on interval $[l,u]$ with respect to the Lebesgue measure (e.g., uniform distribution) then given that the distribution of $\vx$ is also absolutely continuous with respect to the Lebesgue measure on $\reals^m$, the matrix
\[
Z_{m+1}=\begin{pmatrix}
p_1&~~&\vx_1^{\trans}\\
\vdots&~~&\vdots\\
p_{m+1}&~~&\vx_{m+1}^{\trans}
\end{pmatrix}
\]
will be invertible with probability $1$. We defer to \citep{billingsley1979probability} for further details on this.

%

\section{Analysis of the Regret}\label{sec:regret_analysis}

In this section we start by providing a lower bound for the cumulative regret of any policy for the dynamic pricing problem with demand covariates. Then in Section \ref{sec:greedy_optimal} we prove an upper bound for the greedy policy when the seller has access to demand covariates. Finally, in Section \ref{sec:astrology} we will prove that in the case of dynamic pricing without demand covariates, the greedy policy can still be optimal if the firm uses some demand covariates (even though the covariates provide no information). While our proof of the upper bound which is the main technical contribution of this paper does not require any additional assumption on distribution of the demand shocks beyond having zero mean and finite variance, our proof of the lower bound which relies on proof of \cite{keskin2014dynamic} for the lower bound of any policy requires demand shocks to have a Gaussian distribution.

\subsection{Lower Bound Analysis}
\label{sec:lower_bound}

If there is no demand shock in any period, then we can easily learn the true parameters by solving a system of linear equations. However, these shocks have a high impact on learning and consequently on the regret. We can statistically estimate the true parameters, but the estimation error  will lead to an inevitably growing revenue loss (regret) over time. Formally, we can prove that the regret of any policy has a lower bound that grows with time.
\begin{theorem}[Lower bound on regret]
\label{thm:lower}
\sl
Assume that the demand shocks are i.i.d. with distribution $\textrm{N}(0,\sigma_\myep^2)$.
There exists a finite positive constant $c$ such that $\Delta^\pi (T; X_{1:T}) \ge c \log(T)$ for any pricing policy $\pi$, any realized demand covariates $X_{1:T}$, and time horizon $T \ge 3$.
\end{theorem}
For any realized demand covariates $X_{1:T}$, Theorem \ref{thm:lower} gives us a lower bound on the worst-case $T$-period regret of any given policy $\pi$. Therefore, if a policy $\pi$ achieves a regret of order $\log(T)$, the same order as the lower bound, then we call the policy an \emph{asymptotically optimal} policy.
\begin{remark} Authors of \cite{besbes2009dynamic,besbes2011minimax,broder2012dynamic,keskin2014dynamic} also provide lower bounds for the regret in similar problems. However, in their settings, firm has no access to the demand covariates. Theorem \ref{thm:lower} is a non-trivial extension since it shows that increasing the firm's information, via adding the demand covariates that could potentially expedite the learning, does not improve the lower bound.
\end{remark}

\subsection{Greedy Policy is Asymptotically Optimal}
\label{sec:greedy_optimal}

In this section we focus on proving an upper bound for GILS policy.
First we introduce a new notation. For a pricing policy $\pi$, the expected regret over the uncertainty of demand covariates is defined by
\begin{eqnarray}
\Delta_\theta^{\pi}(T) \equiv \E_{\vx} \left[ \Delta_\theta^{\pi}(T; X_{1:T})\right]\,.
\end{eqnarray}
Next, we state the main result of Section \ref{sec:greedy_optimal}. Its proof is provided in Appendix \ref{app:subsec:upperbound4m=1}.
\begin{theorem}[GILS is asymptotically optimal]\label{thm:greedy}
Consider the dynamic pricing problem with demand covariates of Section \ref{sec:problem_formulation} and assume that $\theta\in\Theta$. The GILS policy $\pi^g$ achieves asymptotically optimal expected regret. In particular, there is a constant $C$ such that $\Delta_\theta^{\pi^g}(T) \le C \log(T)$, for all $T$ large enough. In addition, the constant $C$ is equal to
\begin{equation}\label{eq:constantC}
\frac{4|b_{\min}|\,K_0\,\sigma_{\myep}^2}{\lambda_0^2}\,\left[\frac{p_0^2}{2} + \frac{a'^2 + m\,r_{\max}^2 \,x_{\max}^2}{b_{\max}^2}+m \right]\,,
\end{equation}
where
\[
K_0\equiv\frac{a'^2+(r_{\max}^2+b_{\min}^2)\lmax(\Sigma_{\vx})}{4b_{\max}^4}
\]
and
\[
\lambda_0\equiv\min\left[\frac{\delta_0^2}{2},
\frac{\delta_0^2b_{\max}^2}{r_{\max}^2},\frac{\lmin(\Sigma_{\vx})}{2}\right]\,.
\]
\end{theorem}
Theorem \ref{thm:greedy} shows that the incomplete learning problem of GILS is fixed under the disturbance created by the demand covariates, justifying its popularity in practice.
\begin{remark}\label{rem:constantCisFinite}
Looking at the expression \eqref{eq:constantC} for the constant $C$, we highlight that the constant $C$ stays finite even if $r_{\max}=0$ (or $\gamma=0$) which is when the demand covariate carries no information. This has an interesting implication that is studied in Section \ref{sec:astrology}.
\end{remark}

\subsection{Astrology Reports Assist Greedy}
\label{sec:astrology}

One of the most interesting findings of this paper is that Theorem \ref{thm:greedy} can be applied to a special case of the model when $\gamma=0$, that is $D_i = a' + \beta (p_i -p_0) + 0\cdot \vx_i + \myep_i$ for $i = 1, 2, \dots, T$.
\begin{corollary}\label{cor4}
Under the same assumptions as in Theorem \ref{thm:greedy}, the GILS policy $\pi^g$ achieves asymptotically optimal expected regret, that is $\Delta_\theta^{\pi^g}(T) = O(\log(T))$, even when demand covariates have no information (i.e., when $\gamma = 0$).
\end{corollary}
In this situation, the demand covariates are not predictive of the demand. Thus, the optimal price $p^*(\theta,x_i) = a'/(-2\beta)$ is the same optimal price as in the case of not having any demand covariates. However, the presence of the demand covariates in the estimation procedure fundamentally impacts performance of GILS policy. Without the demand covariates, GILS policy has a regret which grows linearly with the total periods. With the uninformative demand covariates, GILS policy achieves the asymptotically optimal performance.

The above analysis demonstrates an interesting approach to the original dynamic pricing problem -- when there is no demand covariate in the demand function. The firm can keep using GILS policy and add some potential demand covariates, as long as they contain random fluctuations, even if they are completely uninformative. The key point is that the firm assumes the covariates could possibly have a predictive power, and uses the covariates to predict the demand function. Then over time, the greedy policy automatically screens out the useless covariates as their estimated coefficients will converge to zero. However, since the coefficients are not exactly zero, the random perturbations provided by them leads to a natural price experimentation which will rescue GILS from incomplete learning. This is why we describe the scenario as ``astrology reports'' help greedy achieve the asymptotic optimal performance.

%

\section{Empirical Analysis}\label{sec:simulations}

In this section, we provide three numerical illustrations of the GILS policy with demand covariates. First, in Section \ref{simu:syn} we demonstrate, on synthetic data, performance and dynamics of GILS with demand covariates. Then in Section \ref{sec:sim-astrology} we consider the case of dynamic pricing without covariates and demonstrate the value of irrelevant covariates (i.e., the astrology reports setting) using synthetic data. Finally, as a robustness check we look at performance of GILS on real data from hotel bookings were the assumptions on distribution demand covariates fail.

\subsection{Performance of GILS with demand covariates}\label{simu:syn}


We generate data by considering the following parameters:
$a' = 0.6$, $\beta = -0.5$, $[b_{\min}, b_{\max}] = [-0.55, -0.4]$, $\sigma_{\myep}=0.05$, $p_0 = 1$, and $[l,u] = [0.75,2]$. For the demand covariates, we assume there are 10 relevant covariates ($m=10$), and each of them is independently drawn from uniform distribution on interval $[-x_{\max},~x_{\max}]$ where $x_{\max}=1.1447$. The true coefficient of demand covariates is taken to be $\gamma = [0.01, 0.01, \cdots, 0.01]^{\trans}\in\reals^{10}$. This means the total variance of all demand covariates would be $\|\gamma\|^2=0.001$.
Finally, we set the bound of the parameter $\gamma$ as $\rmax =1$. Recall that $\rmax$ will be used in truncating $\hat{\theta}$'s.

We simulate GILS policy for $T=10^6$ periods and record the results. Then we repeat the simulation 50 times and present the average performance as well as its 95\% confidence interval, shown via a shaded region, for several metrics as described below.
\begin{figure}[thb]
  \centering
  \includegraphics[width=0.5\textwidth]{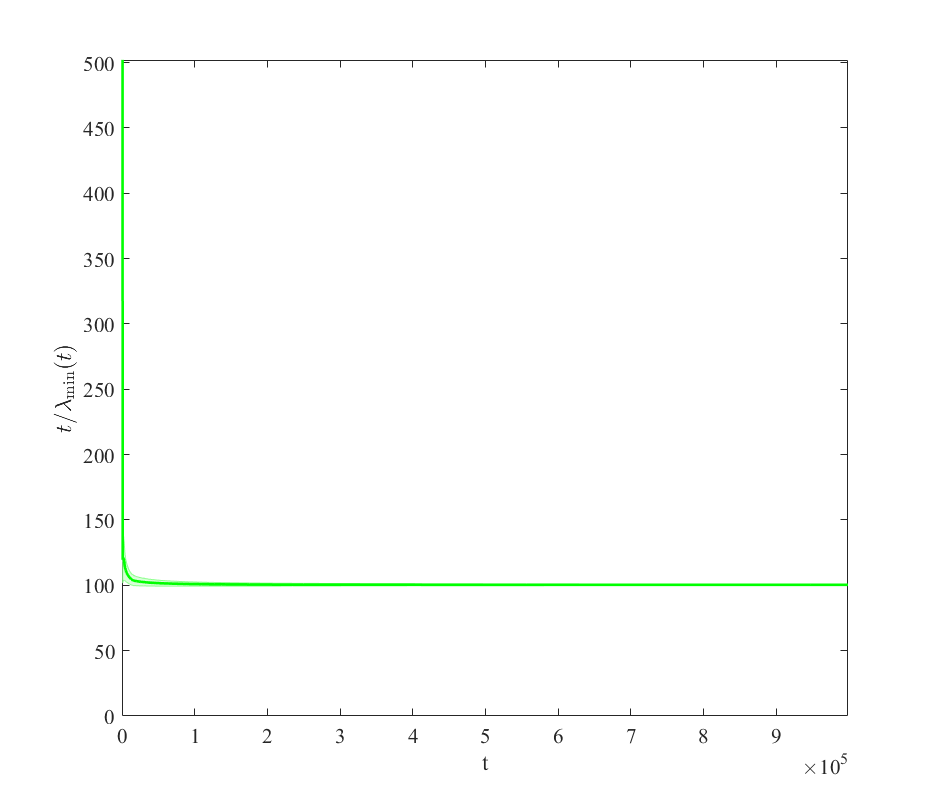}
  \caption{{\bf Dynamics of $\,t/\lambda_{\min}(t)$.} Total periods $T$ is $10^6$, and the problem parameters are $a'=0.6$, $\beta = -0.5$, $p_0 = 1$, $m=10$, $\gamma = [0.01, \cdots, 0.01]^{\trans}\in\reals^{10}$, $[l,u] = [0.75,2]$, $[b_{\min}, b_{\max}] = [-0.55, -0.4]$, $\rmax = 1$, $\sigma_{\myep}=0.05$. }
  \label{fig:n/lambda-min}
\end{figure}
Figure \ref{fig:n/lambda-min} shows the re-scaled minimum eigenvalue of the matrix $Z_t^{\trans} Z_t$ over the total horizon. 
If $t$ refers to the $t^{th}$ time period ($1\le t\le T$), the figure shows that $t/\lambda_{\min}(t)$ is asymptotically bounded near a constant ($100$ here). This means that $\lambda_{\min}(t)$ grows linearly with time periods. Proving this fact rigorously is the main part of our proofs for Theorems \ref{thm:greedy}.

\begin{figure}[thb]
\begin{subfigure}{.5\textwidth}
  \centering
  \includegraphics[width=\textwidth]{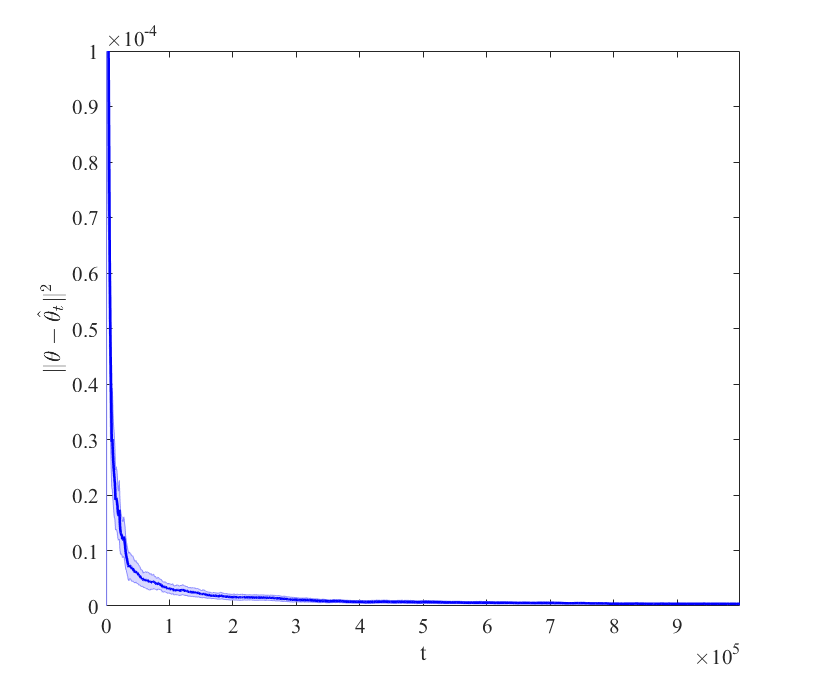}
  \caption{$\|\theta - \hat{\theta}(t)\|^2$ versus $t$.}
\end{subfigure}
\begin{subfigure}{.5\textwidth}
  \centering
  \includegraphics[width=\textwidth]{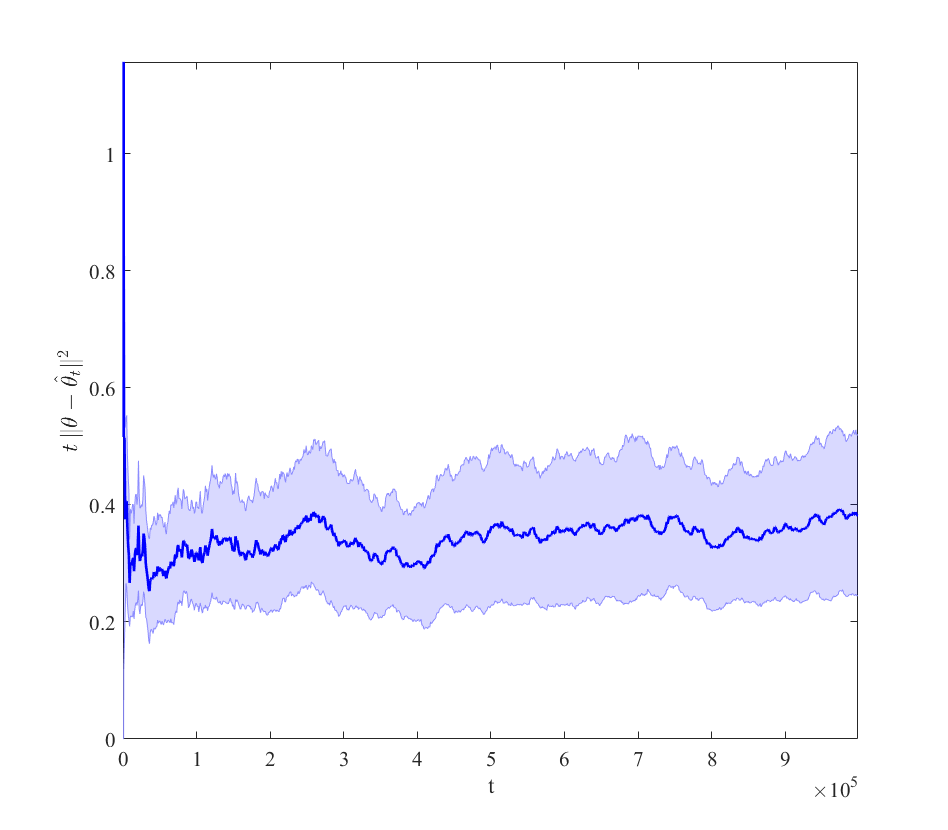}
  \caption{$ t \cdot \|\theta - \hat{\theta}(t)\|^2$ versus $t$.}
\end{subfigure}
\caption{{\bf Estimation error.} Total periods $T$ is $10^6$, and the problem parameters are $a'=0.6$, $\beta = -0.5$, $p_0 = 1$, $m=10$, $\gamma = [0.01, \cdots, 0.01]^{\trans}\in\reals^{10}$, $[l,u] = [0.75,2]$, $[b_{\min}, b_{\max}] = [-0.55, -0.4]$, $\rmax = 0.3162$, $\sigma_{\myep}=0.05$.}\label{fig:estimation-error}
\end{figure}
Figure \ref{fig:estimation-error} shows the estimation error of the parameter vector $\theta$. The left plot shows that the error converges to zero. We re-scaled the error by the time period parameter $t$ in the right plot, and $t \cdot \|\theta - \hat{\theta}(t)\|^2$ fluctuates but stays bounded which means that $\|\theta - \hat{\theta}(t)\|^2$ converges to zero with the rate $\frac{\textrm{constant}}{t}$.

\begin{figure}[thb]
\begin{subfigure}{.5\textwidth}
  \centering
  \includegraphics[width=\textwidth]{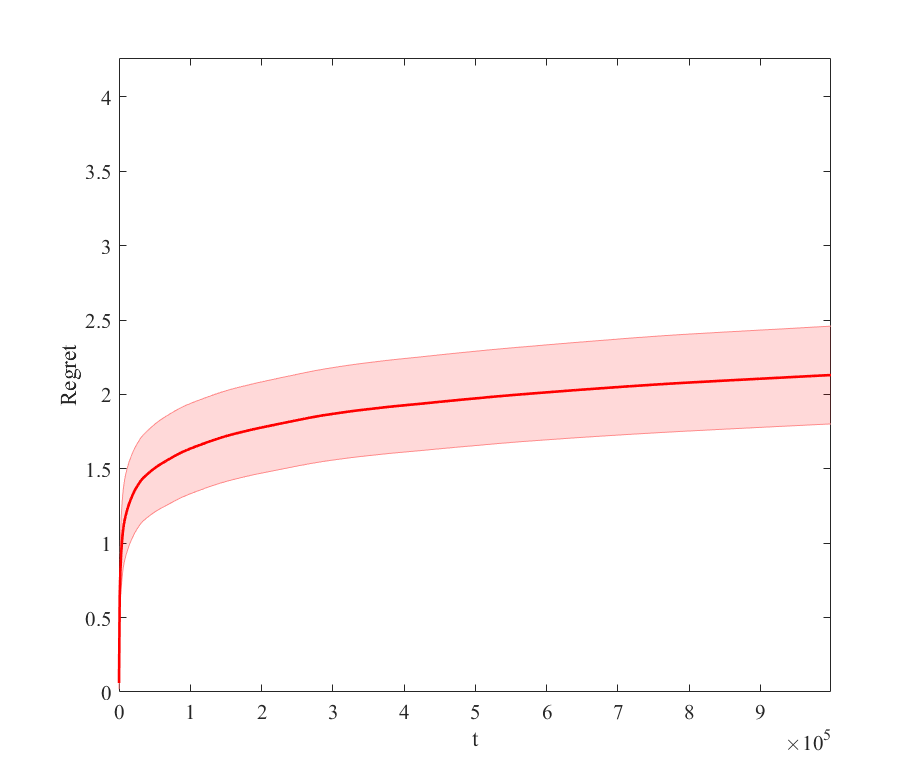}
  \caption{Regret versus $t$.}
\end{subfigure}
\begin{subfigure}{.5\textwidth}
  \centering
  \includegraphics[width=\textwidth]{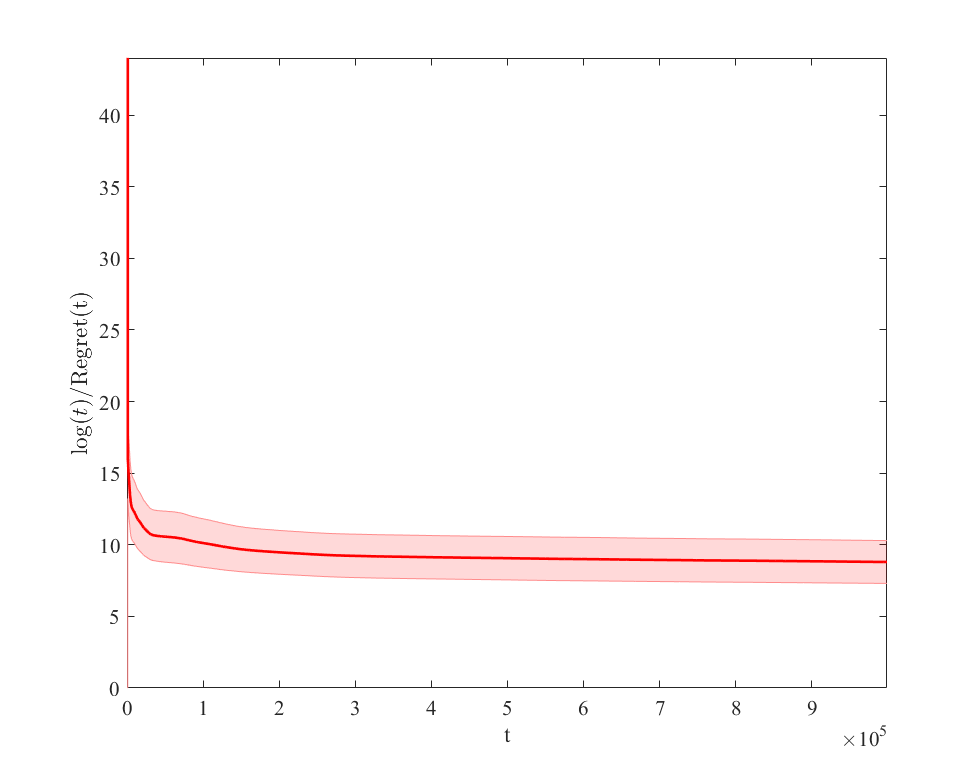}
  \caption{$\log(t)$/regret($t$) versus $t$.}
\end{subfigure}
\caption{{\bf Regret.} Total periods $T$ is $10^6$, and the problem parameters are $a'=0.6$, $\beta = -0.5$, $p_0 = 1$, $m=10$, $\gamma = [0.01, \cdots, 0.01]^{\trans}\in\reals^{10}$, $[l,u] = [0.75,2]$, $[b_{\min}, b_{\max}] = [-0.55, -0.4]$, $\rmax = 1$, $\sigma_{\myep}=0.05$.}\label{fig:regret}
\end{figure}
Figure \ref{fig:regret} shows dynamics of the accumulated regret over the periods. The left plot shows the logarithmic growth of the regret. To see this better, re-scaling the regret to $\log(t)$/Regret($t$) in the right plot, we see that the curve has a positive lower bounded which means the accumulated regret has a growth of order $\log(t)$.

\begin{figure}[thb]
  \centering
  \includegraphics[width=0.5\textwidth]{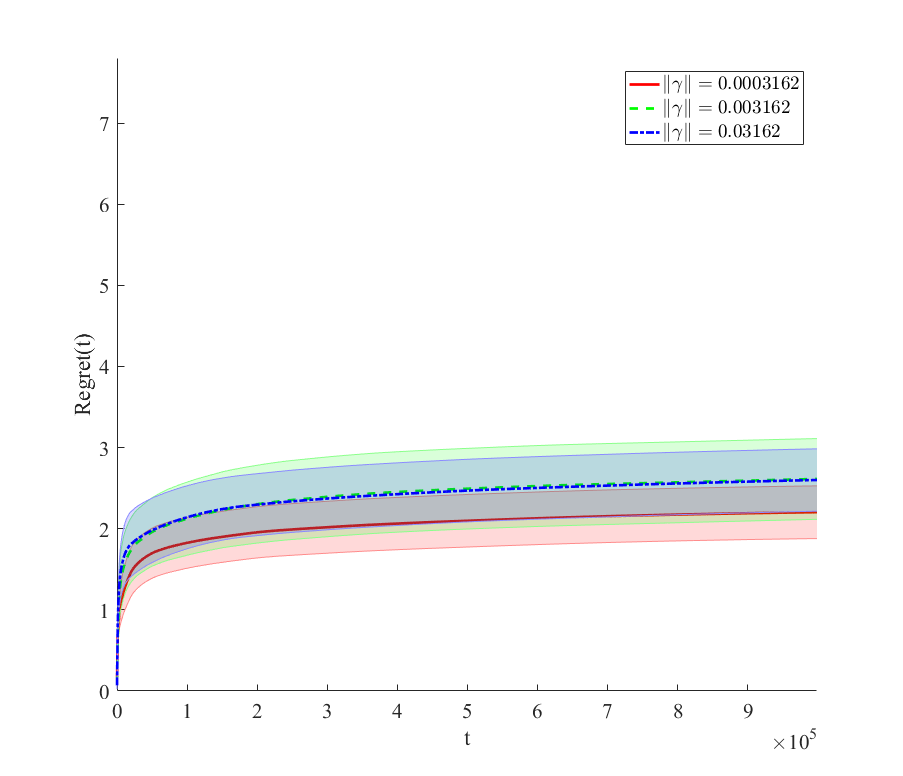}
  \caption{{\bf Regret of GILS with varying $\|\gamma\|_\infty$.} Total period is $T=10^6$, and the problem parameters are $a'=0.6$, $\beta = -0.5$, $p_0 = 1$, $m=10$, $[l,u] = [0.75,2]$, $[b_{\min}, b_{\max}] = [-0.55, -0.4]$, $r_{\max}=1$ $\sigma_{\myep}=0.05$.}\label{fig:regret-vary-gamma}
\end{figure}
Figure \ref{fig:regret-vary-gamma} shows regret of GILS when the effect of demand covariates is varied. In particular, by choosing the parameter $\gamma \in\reals^{10}$ from the set
\[
\left\{[0.0001, \cdots, 0.0001]^{\trans},[0.001, \cdots, 0.001]^{\trans}, [0.01, \cdots, 0.01]^{\trans}\right\},
\]
and keeping all the other parameters as before. We see that when the demand covariates have higher predictive power (larger $\|\gamma\|$) the performance is better.

\subsection{Performance of GILS with astrology reports }\label{sec:sim-astrology}

In this section, we consider the dynamic pricing problem (with incumbent price) studied by \cite{keskin2014dynamic} when there are no demand covariates. We will empirically show that ``astrology reports'' (irrelevant demand covariates) fix the incomplete learning problem of GILS. We also simulate the base version of GILS (the one without covariates) for comparison. To avoid confusion, we denote GILS with astrology report by GILS+.

\begin{figure}[thb]
\begin{subfigure}{.5\textwidth}
  \centering
  \includegraphics[width=\textwidth]{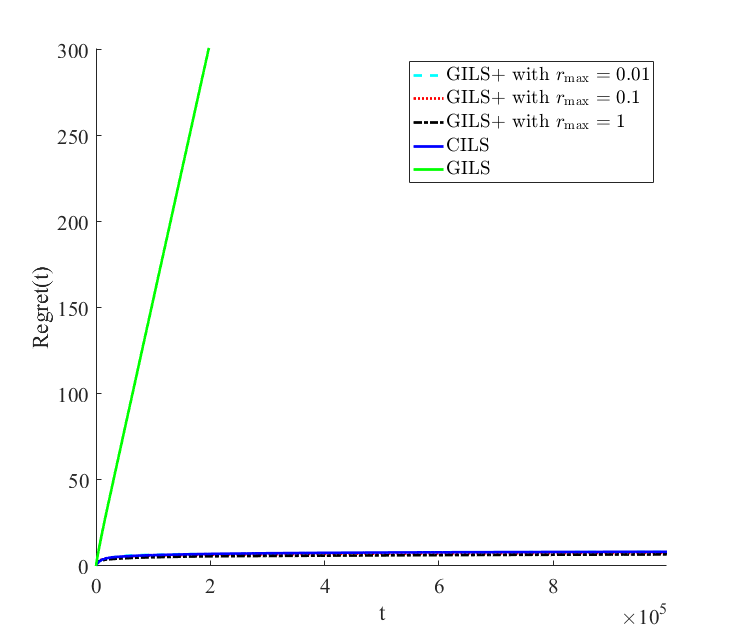}
  \caption{}
\end{subfigure}
\begin{subfigure}{.5\textwidth}
  \centering
  \includegraphics[width=\textwidth]{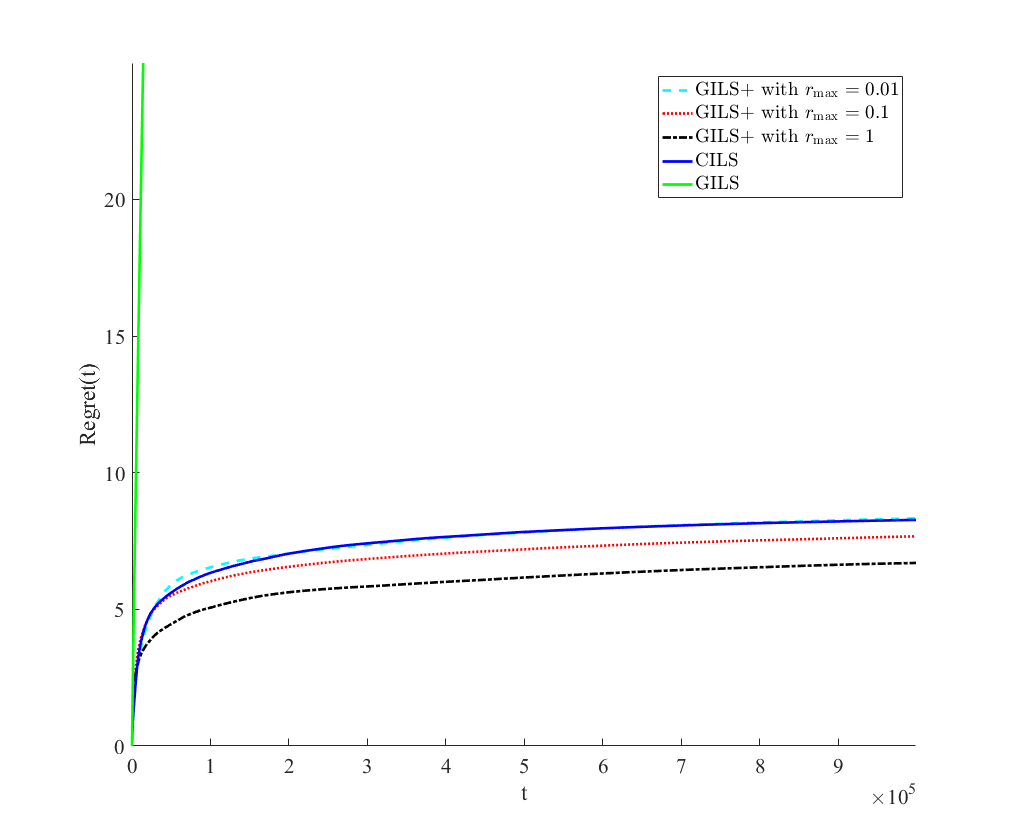}
  \caption{}
\end{subfigure}
\begin{center}
\begin{subfigure}{.5\textwidth}
  \centering
  \includegraphics[width=\textwidth]{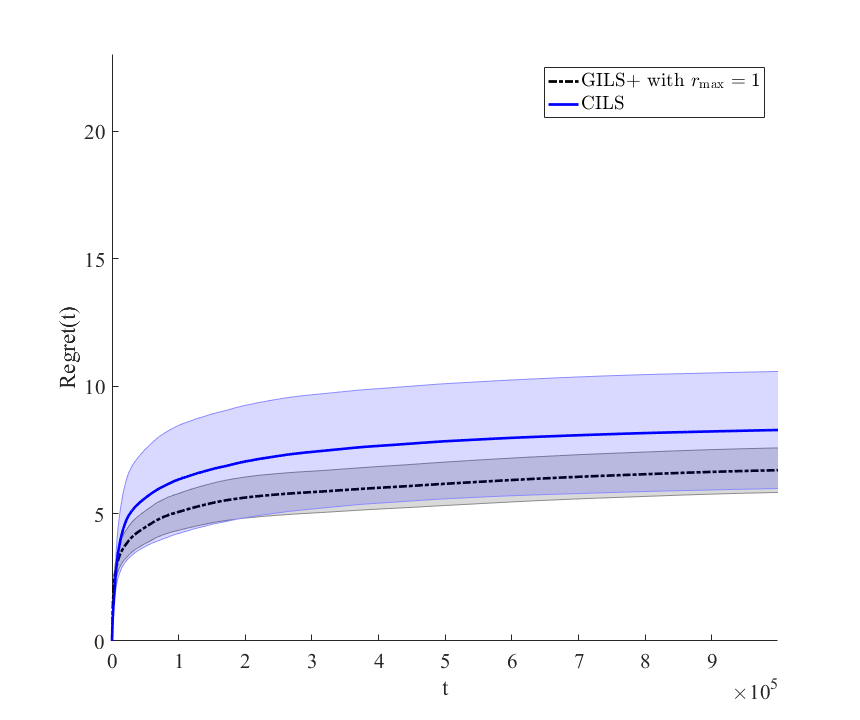}
  \caption{}
\end{subfigure}
\end{center}
\caption{{\bf Regret: GILS+, GILS and CILS.} Total period is $T=10^6$, and the problem parameters are $a'=0.6,\ \beta = -0.5,\ p_0 = 1,\ \gamma = 0,\ \sigma_{\myep} = 0.1,\ [l,u] = [0.75,2],\ [b_{\min}, b_{\max}] = [-0.55, -0.4]$.}\label{fig:astroglogy-regrets-1}
\end{figure}

We use the same parameters as in \citep{keskin2014dynamic} that is
$a' = 0.6$, $\beta = −0.5$, $p_0 = 1$, $[l,u] = [0.75,2]$, $\sigma_{\myep}=0.1$, and the parameter $\kappa$ of CILS is set to $0.1$. We consider a single demand covariate (astrology report variable), drawn randomly from the same uniform distribution as in Section \ref{simu:syn} and varying $r_{\max}\in \{1, 0.1, 0.01\}$. Each algorithm is simulated 50 times up to time $T=10^6$ and average regret (across 50 simulations) is plotted versus time. The results are shown in Figure \ref{fig:astroglogy-regrets-1}. Note that in subfigures (a) and (b) we do not show 95\% confidence regions to reduce the clutter on the graphs since most algorithms have very comparable performance. Figures \ref{fig:astroglogy-regrets-1}(a)-(b) represent the same result with two different scaling of the $y$-axis. From Figure \ref{fig:astroglogy-regrets-1}(a) it is clear that GILS is the worst policy with a regret that grows linearly in time.  After re-scaling the $y$-axis, in Figure \ref{fig:astroglogy-regrets-1}(b), we can see that GILS+ has a fundamentally different behavior than GILS, irrespective of the order of $r_{\max}$ and it seem to marginally outperform CILS. However, in Figure \ref{fig:astroglogy-regrets-1}(c), comparing CILS and even the best GILS+ by depicting the 95\% confidence region across the 50 simulations we see that the observed difference is not statistically significant.

\subsection{Results on Hotel Bookings Data}\label{sec:sim-realdata}

The objective of this section is to test robustness of our main result, demand covariates fix the incomplete learning problem of GILS, by using real data from hotel bookings on Expedia (a large online travel agency). Although we are not considering some important aspects of the real-life dynamic pricing such as inventory constraints, our goal is to provide complementary simulations as in Section \ref{simu:syn} where most parameters of the model are not synthetic and the distributional assumptions on demand covariates may fail.

Our data is from Expedia, Inc. which is an American-based parent company to several global online travel brands including {\tt Expedia.com}, {\tt Hotels.com}, {\tt Hotwire.com}, {\tt Orbitz.com}, and a several more travel websites. The data contains user searches and booking records for 65,000 hotels during November, 2012 to July, 2013. The data is publicly available\footnote{\href{https://www.kaggle.com/c/expedia-personalized-sort}
{https://www.kaggle.com/c/expedia-personalized-sort}} and in addition to search and booking records, contains demand covariates such as price competitiveness, hotel characteristics, location attractiveness, aggregated statistics on purchase history of users, etc. We limited ourselves to
search results that satisfied the following constraints:
\begin{itemize}
\item User looked for one room only for one or two adults (no children).

\item The length of stay is one night.

\item The price per night for the hotel is below \$1000.

\item The search occurred within 7 days of the booking date.
\end{itemize}
These conditions provide a more homogeneous data set by removing search results of strategic customers, complications around multi-night discounts, special requirements for travelers with a large family, or extremely expensive bookings, such that the final data matches better with our dynamic pricing problem.

We also aggregate the search results and bookings for each (day, hotel) pair as follows. The total number of booked rooms for each hotel on that day is taken as a proxy for the demand and the average displayed price is taken as a proxy for the price. We also select several demand covariates from the data. They are the \textsf{Star} rating of the hotel, the average \textsf{Review} score of the property, whether the hotel is part of a major \textsf{Brand} chain, hotel's {\sf Position} on search results page, whether the check-in day is a {\sf Weekend}, a score for the desirability of a hotel's {\sf Location} and whether the booking date is in {\sf Summer} season. This led to our final data set with $267,832$ rows of (day,hotel) pairs. Table \ref{tab:expedia-data} demonstrates samples from the final data set with the demand covariates. However, for our simulation we will standardize all covariates to have mean a $0$ and and variance of $1$.

\begin{table}[h]
\centering \small
\begin{tabular}{ccccccccc}
\toprule
& & \multicolumn{7}{c}{Demand Covariates} \\
    \cmidrule(r){3-9}
{\sf Demand}	& {\sf Price (\$)}  &	{\sf Star}	&	{\sf Review }	&	{\sf Brand	}&	{\sf Position} &	{\sf Weekend } &	{\sf Location}	&	{\sf Summer	}\\
\midrule
2	&	144	&	4	&	3.5	&	1	&	15	&	0	&	0.0058	&	1	\\
2	&	95	&	3	&	4	&	1	&	29	&	0	&	0.0157	&	1	\\
3	&	182.16&	4	&	4.5	&	1	&	14	&	0	&	0.28724	&	0	\\
3	&	68	&	2	&	4	&	0	&	16.3&	1	&	0.0667	&	0	\\
3	&	154	&	4	&	4	&	1	&	9.75&	1	&	0.1249	&	1	\\
0	&	146.3&	3	&	4	&	0	&	15.25&	0	&	0.0414	&	1	\\
\vdots	&	\vdots	&	\vdots	&	\vdots	&	\vdots	&	\vdots	&	\vdots	&	\vdots	&	\vdots	\\
\bottomrule
\end{tabular}
\caption{Sample Rows of the Processed Hotel Bookings Data}
\label{tab:expedia-data}
\end{table}

The data also contains average historical price for all these hotels prior to November, 2012 ($\$129.92$) that we use as incumbent price. That is $p_0=\$129.92$.

To obtain the true parameters of the demand model, known to a clairvoyant seller, we use a linear regression on all $267,832$ rows. The result is shown below.
\begin{eqnarray}\label{eq:true-regression}
\mathsf{Demand} &=& 0.06361 -0.0001192 \cdot \mathsf{Price} + 0.007164 \cdot \mathsf{Star}
+ 0.00292\cdot \mathsf{Review} \\
&& -0.0008887\cdot \mathsf{Brand} -0.02568\cdot \mathsf{Position} + 0.001736\cdot \mathsf{Weekend} \nonumber\\
&& + 0.01519\cdot \mathsf{Location} + 0.001228 \cdot \mathsf{Summer}\nonumber\,.
\end{eqnarray}
where all coefficients are statistically significant with p values less than $3\%$.
After estimating the true parameters, in order to simulate a pricing policy, we
will ignore the columns {\sf Demand} and {\sf Price} and only focus on the (standardized) demand covariates. For each of the 50 simulation runs, we randomly permute rows of the data and assume they are sequentially realized over time. Then GILS will dynamically adjust the price according to the demand covariates, and estimates the parameter vector $\theta$. At each time period, we use the true parameters from Eq. \eqref{eq:true-regression} to calculate the realized demand for the price (suggested by GILS). Similarly, we can calculate the optimal price for that realization of demand covariates via the true parameters and hence can calculate the regret of that time period.

Finally, we measure the accumulated regret up to time period $T=267,832$ that all samples are observed. We repeat the simulation 50 times (each time with a different random permutation of the rows) to obtain the average and a 95\% confidence region around the average for the $t/\lmin(t)$, $\log(t)/\textrm{Regret}(t)$, and $t\|\theta-\hat{\theta}_t\|^2$. We note that since a random permutation of the same data set is used each time, the 50 samples are not independent and hence the provided confidence regions are not correct 95\% intervals. We only plot them as a rough measure of statistical fluctuations. Other parameters in the model are $\rmax = 1$, $b_{\min} = -1^{10}$ and $b_{\max} = -1^{-10}$. Note that these values are selected very conservatively which leads to a very large range of parameters $\Theta$. This choice is done to demonstrate robustness of the results, but in practice one can use domain knowledge or prior data to choose a smaller region.

\begin{figure}[thb]
  \begin{subfigure}{.5\textwidth}
    \centering
    \includegraphics[width=\textwidth]{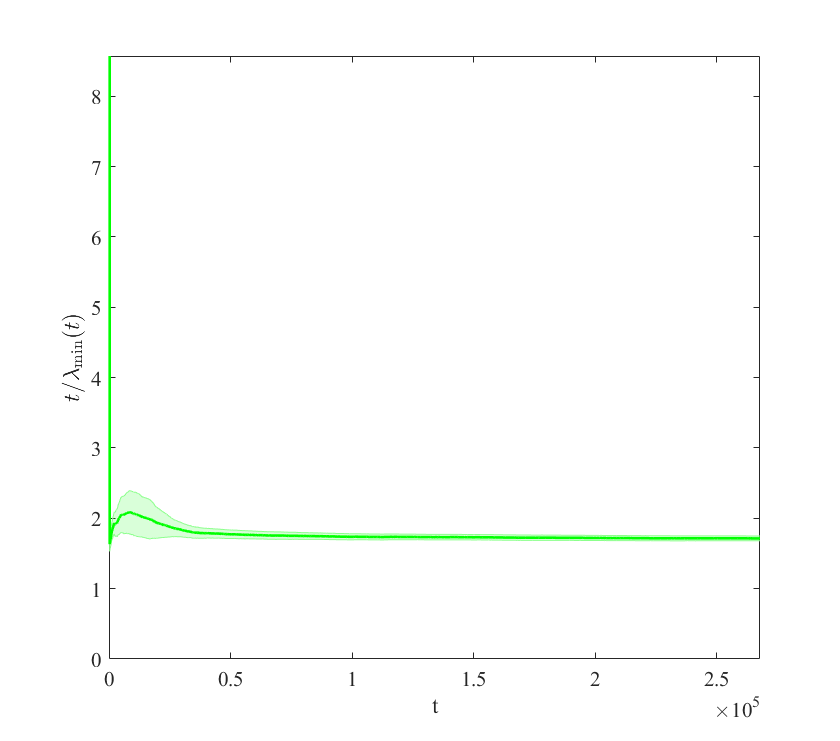}
    \caption{}
  \end{subfigure}
  \begin{subfigure}{.5\textwidth}
    \centering
    \includegraphics[width=\textwidth]{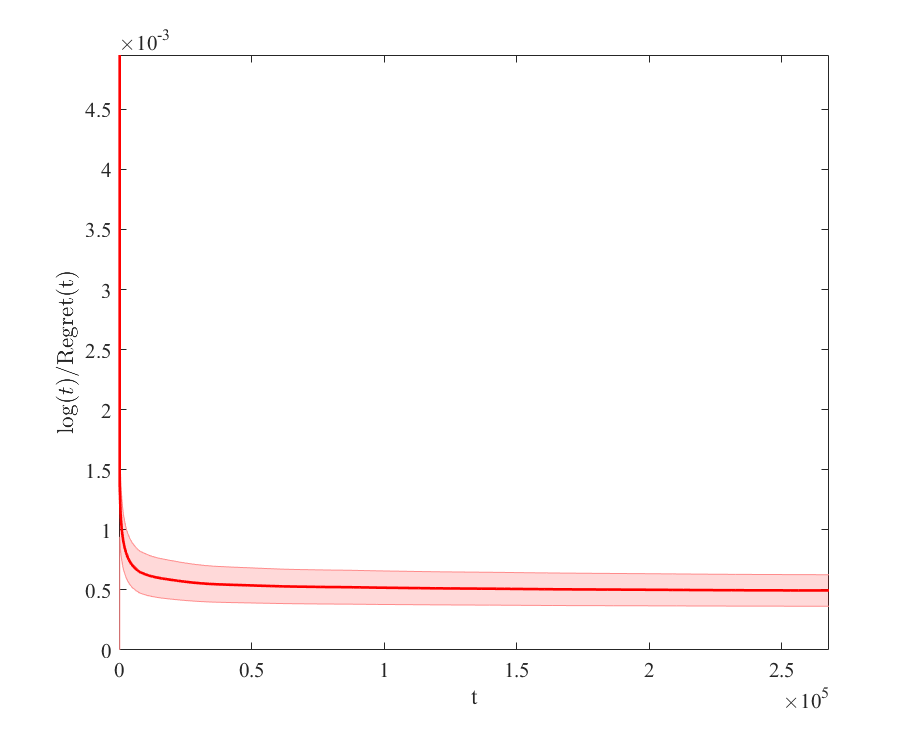}
    \caption{}
  \end{subfigure}
  \begin{center}
  \begin{subfigure}{.5\textwidth}
    \centering
    \includegraphics[width=\textwidth]{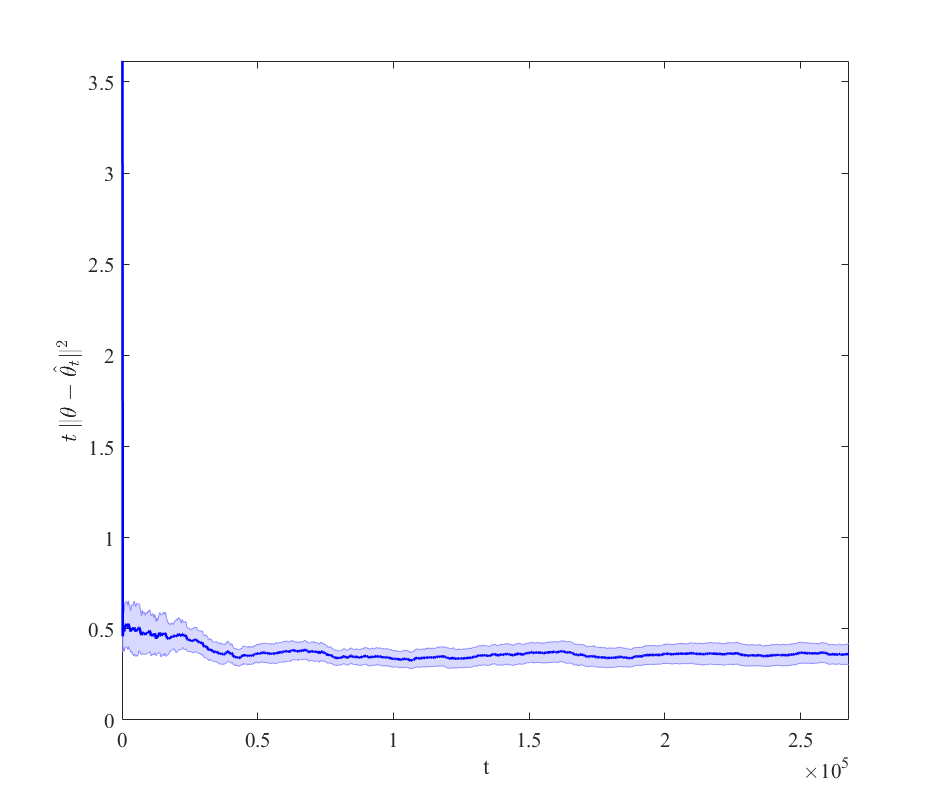}
    \caption{}
  \end{subfigure}
  \end{center}
  \caption{{\bf Performance of GILS on Expedia data.} The total number of time periods is 267,832.}\label{fig:GILS-Expedia}
\end{figure}
Results are shown in Figure \ref{fig:GILS-Expedia}; $(a)$ shows that $t/\lmin(t)$ has a constant upper bound, (b) illustrates that $\log(t)/\textrm{Regret}(t)$ is asymptotically strictly positive i.e., the regret grows by a constant times $\log(t)$, and (c) shows that $t\|\theta-\hat{\theta}_t\|^2$ has a constant upper bound. All of these are in-sync with simulation results of Section \ref{simu:syn} as well as our theoretical results even though most of the assumptions of the model are not satisfied. For example, the assumption on distribution of demand covariates is not guaranteed or condition \eqref{eq:delta-condition} does not hold. This demonstrates robustness of the results.

%

\section{Conclusions and Discussion}\label{sec:discussion}

This paper considers a dynamic pricing problem when the demand function is unknown. The firm leverages extra information -- demand covariates -- to determine the price in each period, in order to maximize cumulative revenue over time. A popular approach in practice for this problem is GILS that focuses only on mazimizing the revenue and ignores learning of the demand function. But it has been shown theoretically and empirically that GILS has a suboptimal performance when the demand function does not contain demand covariates.
In this paper, we proved that GILS is asymptotically optimal when there are demand covariates. This result holds even when the predictive power of the demand covariates becomes negligible. An interesting corollary of this result is that, when the demand covariates have no predictive power, including them in GILS will fix the incomplete learning problem of GILS and help it achieve the asymptotically optimal performance. The results are shown to be robust using simulations on synthetic and real data sets, in particular in situations where our modeling assumptions fail.

The phenomena that demand covariates assist GILS to achieve the asymptotic optimal performance can be extended in several ways. First, while the linear demand model is well accepted in the economics literature and revenue management practice, it is too simple and restrictive. Therefore, extending the results to the cases with a more complex demand model with covariates is a promising future research direction.
Examples would include generalized linear demand models, as studied by \cite{den2014simultaneously}, or models with more general dependence on demand covariates. Second, this paper focuses on the simple case that firm only sells one type of products. We expect generalizations to the cases with multiple products, as in \citep{keskin2014dynamic,den2014dynamic}, to be valid and plan to pursue them. Third, the interesting phenomena that disturbances obtained from the demand covariates save the greedy policy from converging to a local (wrong) optimal point, is an intriguing direction to study in other sequential decision making problems.

\ACKNOWLEDGMENT{The authors are grateful to Michael J. Harrison
for encouragements and insightful discussions throughout this study and to Yonatan Gur, Dan Iancu, and Stefanos Zenios for guidance and suggestions.

In addition, M. Bayati acknowledges the support of NSF awards CCF:1216011 and CMMI:1451037.}

%
%
%

%

 \begin{APPENDICES}

\section{Proofs}\label{sec:proofs}

 \subsection{Proof of Theorem \ref{thm:lower}}

 \proof{}  To obtain the lower bound on the regret, we build upon the result of \cite{keskin2014dynamic}.
Our argument is by contradiction. Assume the contrary, that there is a realized sequence of demand covariates $X_{1:T}$ and a special policy $\tilde{\pi}$ where for each $t$, $\tilde{\pi}_t:\tilde{H}_t\rightarrow [l,u]$ with $\tilde{H}_t=(D_1,\dots,D_{t-1},p_1,\dots,p_{t-1},\vx_1,\dots,\vx_t)$, and $\tilde{\pi}$ is such that for any $c>0$ there exist a $T_c\ge 3$ with
 \begin{eqnarray}\label{ieq:contr}
 \Delta^{\tilde{\pi}} (T_c; X_{1:T_c}) < c \log(T_c)\,.
 \end{eqnarray}
 %
Now using the fact that $\Delta^{\tilde{\pi}} (T; X_{1:T}) = \sup_{\theta\in\Theta} \Delta^{\tilde{\pi}}_\theta (T; X_{1:T})$, the inequality (\ref{ieq:contr}) would mean that for all $\theta \in \Theta$,
 \begin{eqnarray}
 \Delta_\theta^{\tilde{\pi}} (T_c; X_{1:T_c})< c \log(T_c)\,.
 \end{eqnarray}
 Now, we consider a very special case of $\theta$, i.e., $\gamma=0$, and the new parameter is $\theta^*=(\beta,0)$. In this case, the demand does not rely on demand covariates any more, that is,
 \begin{eqnarray*}
 D_s &=& a' + \beta p_s + \gamma\cdot \vx_s + \myep_s = a' + \beta p_s + \myep_s\quad \hbox{ for all } s=1,2,\dots
 \end{eqnarray*}
Therefore, if we consider the dynamic pricing problem without demand covariates and with the unknown parameter $\theta_0 = \beta$ then the policy $\pi_0=\tilde{\pi}$, even though it may use the irrelevant information provided by the covariates, is a feasible policy for this problem. Thus, we have the regret of the dynamic pricing problem without demand covariates satisfying
\begin{eqnarray*}
\Delta^{\pi_0}_{\theta_0} (T_c) < c \log(T_c)\,,
\end{eqnarray*}
for any $c>0$ and any $\theta_0\in\Theta_0$ where $\Theta_0$ is projection of $\Theta$ to its first coordinate (the coordinate corresponding to the coefficient of price). This means we have
\begin{eqnarray*}
\Delta^{\pi_0} (T_c) = \sup_{\theta_0\in\Theta_0}\Delta^{\pi^0}_{\theta_0} (T_c)< c \log(T_c)\,.
\end{eqnarray*}
for any constant $c>0$. But, the above statement contradicts Theorem 3 in \citep{keskin2014dynamic}, where they show that there is a positive constant $c_1$ such that
for all feasible policies $\pi_0$ for the dynamic pricing problem without demand covariates ($\gamma=0$) and for all $T\ge 3$,
\begin{eqnarray}
\Delta^{\pi_0} (T) \ge c_1 \log(T)\,.
\end{eqnarray}
This finishes our proof of Theorem \ref{thm:lower}.

 \endproof

%
%
%
%

\subsection{Proof of Theorem \ref{thm:greedy}}\label{app:subsec:upperbound4m=1}

\proof{}
First, we notice that $\Delta_\theta^\pi (T)$ can be written as below:
\begin{eqnarray}\label{eq:regret}
\Delta_\theta^\pi (T) = -\beta \sum_{t=1}^{T}\E_{\vx} \E_{\myep} \left\{ [p_t - p^*(\theta,\vx_t) ]^2\right\}.
\end{eqnarray}
Here $-\infty<b_{\min}\le \beta\le b_{\max}<0$ is bounded, and recall that the expectations $\E_{\vx}$ and $\E_{\myep}$ are with respect to the randomness of $\{\vx_r\}_{1\le r\le t}$ and $\{\myep_r\}_{1\le r\le t}$ respectively. In order to prove an upper bound on the regret, we provide an upper bound on each term $\E_{\vx} \E_{\myep} \left\{ [p_t - p^*(\theta,\vx_t) ]^2\right\}$ separately, and then sum them up. Also, for simplicity, throughout this proof we will use a single expectation notation $\E$ to refer to both expectations. We also fix the policy to be the greedy policy $\pi^g$ and assume the parameter $\theta$ is fixed.

Second, note that the revenue loss of the greedy policy is due to the least squares estimation errors. To see this we note that the expected value of squared pricing error in period $t$ is given by
\begin{eqnarray}
\E \left\{[p^*(\theta,\vx_{t}) - p_{t}]^2\right\} =
\E \left\{[p^*(\theta,\vx_{t}) - p^*(\vartheta_{t-1},\vx_{t})]^2\right\}
\end{eqnarray}
where $\vartheta_t$ is the truncated least squares estimate calculated at the end of period $t$. Further, denoting the coordinates of $\vartheta_t$ by $(\tilde{\beta}_t, \tilde{\gamma}_t)$, we can bound the expectation (only with respect to $\vx_t$) of squared price deviation by
\begin{align}
\E \Big\{ [ p^*(\theta,\vx_{t}) - p^*(\vartheta_{t-1},\vx_{t}) ]^2 \Big| H_{t-1} \setminus \{\vx_t\} \Big\}&= \E \left\{ \left[ \frac{a' + \gamma^{\trans}\vx_{t}}{-2\beta} - \frac{a' + \tilde{\gamma}_{t-1}^{\trans}\vx_{t}}{-2 \tilde{\beta}_{t-1}}  \right]^2 \Big| H_{t-1} \setminus \{\vx_t\} \right\}  \nonumber\\
&\le \left( \frac{a'}{-2\beta} - \frac{a'}{-2\tilde{\beta}_{t-1}}\right)^2 + \|\xi_{t-1}\|^2\lmax(\Sigma_{\vx})\label{eq:2}\,,
\end{align}
where
$
\xi_{t-1}\equiv \frac{\gamma}{-2\beta} - \frac{\tilde{\gamma}_{t-1}}{-2\tilde{\beta}_{t-1}}$. Also, recall that $H_{t-1} \setminus \{\vx_t\} = \{D_1,\dots,D_{t-1}, p_1,\dots, p_{t-1}, \vx_1,\dots,\vx_{t-1} \}$ is the history $H_{t-1}$ without the current demand covariates $\vx_t$. Thus, $\tilde{\beta}_{t-1}$ and $\tilde{\gamma}_{t-1}$ are determined (measurable) with respect to $H_{t-1} \setminus \{\vx_t\}$, while $\vx_t$ is a random variable with $\E[\vx_t] =0$, $\E[\vx_t^{\trans}\vx_{t}] = \Sigma_{\vx}$. Next, we will find upper bounds for the right hand side of Eq. \eqref{eq:2}.  In particular, using
\begin{align*}
\left( \frac{a'}{-2\beta} - \frac{a'}{-2\tilde{\beta}_{t-1}}\right)^2 &\le \frac{a'^2}{4b_{\max}^4}(\beta-\tilde{\beta}_{t-1})^2
\end{align*}
and
\begin{align*}
\|\xi_{t-1}\|^2 &\le \frac{r_{\max}^2(\beta-\tilde{\beta}_{t-1})^2+b_{\min}^2\|\gamma-\tilde{\gamma}_{t-1}\|^2}{4b_{\max}^4}\,,
\end{align*}
we have
\begin{align}
\E \Big\{ [ p^*(\theta,\vx_{t}) - p^*(\vartheta_{t-1},\vx_{t}) ]^2 \Big| H_{t-1} \setminus \{\vx_t\} \Big\}&\le K_0\|\theta-\vartheta_{t-1}\|^2\,,\label{eq:K0Lip}
\end{align}
where
\[
K_0\equiv\frac{a'^2+(r_{\max}^2+b_{\min}^2)\lmax(\Sigma_{\vx})}{4b_{\max}^4}\,.
\]

Since $\theta \in \Theta$, and $\vartheta_{t-1}$ is projection of
$\widehat{\theta}_{t-1}$ onto $\Theta$, we have $\| \theta - \vartheta_{t-1} \|^2 \le \| \theta - \widehat{\theta}_{t-1} \|^2$.
Therefore, we will try to find an upper bound for $\E [\|\theta - \widehat{\theta}_{t-1} \|^2]$. To do this, we express the estimation error in terms of the error vector $\mathbf{e}_t$ and $Z_t$ according to equation (\ref{eq:dev}). In particular,
\begin{eqnarray}
\| \theta - \widehat{\theta}_t \|^2 &=& \mathbf{e}^{\trans}_t Z_t (Z_t^{\trans} Z_t)^{-2} Z_t^{\trans} \mathbf{e}_t \nonumber\\
&\le & \mathbf{e}^{\trans}_t Z_t \left( \frac{1}{\lmin(t)}\right)^2 Z_t^{\trans} \mathbf{e}_t \nonumber \\
&=& \frac{\mathbf{e}^{\trans}_t Z_t Z_t^{\trans} \mathbf{e}_t}{\lmin^2(t)},
\end{eqnarray}
where $\lmin(t)$ is the minimum eigenvalue of the matrix $Z_t^{\trans} Z_t$. The last inequality is a consequence of
\begin{eqnarray}
Z_t^{\trans} Z_t \succeq \lmin(t) I ~~~~\textrm{and}~~~~ (Z_t^{\trans} Z_t)^{-1} \preceq \frac{1}{\lmin(t)} I\,,
\end{eqnarray}
where matrix inequality $A\succeq B$ means $A-B$ is positive semi-definite.

Since $Z_t =(u_1, u_2, \dots, u_t)^{\trans}$ and $\displaystyle u_i = {p_i - p_0\choose \vx_i}$, we have
\begin{eqnarray}
Z_t^{\trans} Z_t = \sum_{i=1}^t u_i u_i^{\trans} = \sum_{i=1}^t \left[ \begin{matrix}
(p_i-p_0)^2 & (p_i-p_0) \vx_i^{\trans}\\
\vx_i(p_i-p_0) & \vx_i \vx_i^{\trans}
\end{matrix} \right]\,.
\end{eqnarray}

Next, we will prove a lower bound for $\lmin(t)$ using the following variant of matrix Chernoff inequality for adapted sequences from \cite{tropp2011user}.
\begin{theorem}[Theorem 3.1 in \citep{tropp2011user}]\label{thm:thropp}
Let $(\Omega,\cF,\P)$ be a probability space.
Consider a finite sequence $\{\bfX_k\}_{k\ge 1}$ of positive-semidefinite matrices with dimension $d$ that is adapted to a filtration $\{\cF_k\}_{k\ge 1}$, i.e.
\[
\cF_0\subset\cF_1\subset\cdots\subset\cF~~~\textrm{and
$\forall k,~\bfX_k$ is $\cF_k$ measurable}\,.
\]
Now suppose that
\[
\lambda_{\max}(\bfX_k)\le R
\]
almost surely. Then, for all $\mu>0$ and all $\zeta\in[0,1)$,
\[
\mathbb{P}\left\{\lmin\left(\sum_k\bfX_k\right)\le (1-\zeta)\mu~~~\textrm{and}~~~\lmin\left(\sum_k\E[\bfX_k|\cF_{k-1}]\right) \ge \mu \right\}\le d\cdot\left[\frac{e^{-\zeta}}{(1-\zeta)^{1-\zeta}}\right]^{\mu/R}\,.
\]
\end{theorem}

In order to apply Theorem \ref{thm:thropp} we need to define the filtration and the adapted sequence. Let $\cF_1\equiv\emptyset$ and, for $i\ge 1$, let $\cF_i$ be the sigma algebra generated by random variables $ D_1,\dots,D_{i-1}, p_1,\dots, p_{i-1}, \vx_1,\dots,\vx_{i-1}$ and let $\bfX_i\equiv u_iu_i^{\trans}$. Then we have, for all $i$
\begin{eqnarray*}
\E[u_iu_i^{\trans}\mid\cF_{i-1}] &=&
\left[
\begin{matrix}
\left(\frac{a'}{-2\tilde{\beta}_{i-1}} - \frac{p_0}{2} \right)^2 + \frac{\tilde{\gamma}_{i-1}^{\trans}\Sigma_{\vx}\tilde{\gamma}_{i-1}}{4\tilde{\beta}_{i-1}^2} &\qquad \frac{\tilde{\gamma}^{\trans}_{i-1}\Sigma_{\vx}}{-2\tilde{\beta}_{i-1}}\\
&\\
 \frac{\Sigma_{\vx}\tilde{\gamma}_{i-1}}{-2\tilde{\beta}_{i-1}}& \qquad \Sigma_{\vx}
\end{matrix}
\right]\\
&& \\
& = &
\left[
\begin{matrix}
\delta_{i-1}^2 + \vq_{i-1}^{\trans}\Sigma_{\vx}\vq_{i-1} & \quad \vq_{i-1}^{\trans}\Sigma_{\vx}\\
&\\
\Sigma_{\vx}\vq_{i-1} & \quad \Sigma_{\vx}
\end{matrix} \right]\,
\end{eqnarray*}
where we introduced two new variables
 \[
 \delta_{i-1} \equiv \left(\frac{a'}{-2\tilde{\beta}_{i-1}} - \frac{p_0}{2}\right)~~~~,~~~~ \vq_{i-1} \equiv \frac{\tilde{\gamma}_{i-1}}{-2\tilde{\beta}_{i-1}}\,.
 \]
Next we state the following result on the minimum eigenvalue of $\E[u_iu_i^{\trans}\mid\cF_{i-1}] $.
\begin{lemma}\label{lem:lmin_EuuT}
We have $\lmin(\E[u_iu_i^{\trans}\mid\cF_{i-1}])\ge \lambda_0>0$ where
\[
\lambda_0\equiv\min\left[\frac{\delta_0^2}{2},\frac{\delta_0^2b_{\max}^2}{r_{\max}^2},\frac{\lmin(\Sigma_{\vx})}{2}\right]\,.
\]
\end{lemma}
We will finish proof of Theorem \ref{thm:greedy} using Lemma \ref{lem:lmin_EuuT} and prove the lemma afterwards.

Applying Lemma \ref{lem:lmin_EuuT}, we have that
\begin{eqnarray}\label{eq:lam_min_lower}
\lmin\left( \sum_{i=1}^t \E[u_iu_i^{\trans}\mid \cF_{i-1}] \right) \ge \sum_{i=1}^t \lmin\left(  \E [u_iu_i^{\trans}\mid \cF_{i-1}] \right) = \lambda_0 t\,,
\end{eqnarray}
and we note that, the maximum eigenvalue of each matrix $u_iu_i^{\trans}$ is upper bounded by a constant $R$ uniformly,
\begin{eqnarray}
\lambda_{\max}(u_iu_i^{\trans}) &\le& \textrm{Tr}(u_iu_i^{\trans})\nonumber\\
&=&\|\vx_i\|^2 + (p_i -p_0)^2 \le m\,x_{\max}^2 + \left(\frac{a' + \tilde{\gamma}_{i-1}^{\trans} \vx_i}{-2 \tilde{\beta}_{i-1}} -\frac{p_0}{2} \right)^2 \nonumber\\
&\le & m\,x_{\max}^2 + \frac{p_0^2}{2} + \frac{a'^2 + m\,r_{\max}^2\, x_{\max}^2}{b_{\max}^2}\equiv R\,.\label{eq:lam_max_upper}
\end{eqnarray}
Now, combining Eqs. \eqref{eq:lam_min_lower} and \eqref{eq:lam_max_upper} with the fact that each matrix $u_iu_i^{\trans}$ is positive semi-definite we can use Theorem \ref{thm:thropp} with $\zeta=1/2$ and obtain
\begin{eqnarray}
\P\left\{ \lmin(t) \le \frac{\lambda_0 t}{2} \right\}  &=& \P \left\{ \lmin \left(\sum_{i=1}^{t} u_iu_i^{\trans}\right) \le \frac{\lambda_0 t}{2}~~ \hbox{ and }~~ \lmin \left( \sum_{i=1}^T\mathbb{E}[u_iu_i^{\trans}\mid\cF_{i-1}] \right) \ge \lambda_0 t \right\} \nonumber\\
&\le & (m+1)\cdot (2e)^{-\frac{\lambda_0 t}{2R}}\,.
\end{eqnarray}

Now, we have enough tools to finalize proof of Theorem \ref{thm:greedy}.
First we recall from Eq. \eqref{eq:K0Lip} that the expected price deviation at time $t$ is upper bounded by
$K_0\,\E\left[\| \theta - \vartheta_t \|^2\right]$ and that $\| \theta - \vartheta_t \|^2\le \| \theta -  \widehat{\theta}_t\|^2$.
We also use the fact that $\theta$ and $\vartheta_t $ belong to $\Theta$ to get the bound
\[
\| \theta -  \vartheta_t \|^2\le(b_{\max}-b_{\min})^2+4r_{\max}^2\,.
\]
Combining all of these we have,
\begin{eqnarray}\label{eq:1}
\E\left[\| \theta - \vartheta_t \|^2\right]&=&
\E\Big[\| \theta - \vartheta_t \|^2\cdot I(\lmin(t)\ge \lambda_0 t/2)\Big]
+
\E\Big[\| \theta - \vartheta_t \|^2\cdot I(\lmin(t)<\lambda_0 t/2)\Big]\nonumber\\
&\le&
\E\left[\| \theta - \widehat{\theta}_t \|^2\cdot I(\lmin(t)\ge \lambda_0 t/2)\right]
+
\left[(b_{\max}-b_{\min})^2+4r_{\max}^2\right]\P[\lmin(t)<\lambda_0 t/2]\nonumber\\
&\le&
\E \left[ \frac{\mathbf{e}^{\trans}_t Z_t Z_t^{\trans} \mathbf{e}_t}{\lmin^2(t)}\cdot I(\lmin(t)\ge \lambda_0 t/2) \right]
+
(m+1)\left[(b_{\max}-b_{\min})^2+4r_{\max}^2\right](2e)^{-\frac{\lambda_0 t}{2R}}\nonumber\\
&\le&
\frac{\E[\mathbf{e}^{\trans}_t Z_t Z_t^{\trans} \mathbf{e}_t]}{(\lambda_0 t)^2/4}
+
(m+1)\left[(b_{\max}-b_{\min})^2+4r_{\max}^2\right](2e)^{-\frac{\lambda_0 t}{2R}}\,.
\end{eqnarray}

Next, we notice that for any $i\le j$, $\myep_j$ is independent of $\myep_i$, $\vx_i$ and $p_i = p^*(\vartheta_{i-1},\vx_i)$, thus $\myep_j$ is independent of $u_i$. With $Z_t^{\trans} \mathbf{e}_t = \sum_{i=1}^{t} u_i \myep_i$ and $\E[\myep_i] =0$, we have
\begin{eqnarray}
\E [\mathbf{e}^{\trans}_t Z_t Z_t^{\trans} \mathbf{e}_t] &=& \E \left[\sum_{1\le i,j\le t} \myep_j u_j^{\trans} u_i \myep_i \right] \nonumber\\
&=& \E \left[\sum_{1\le i < j\le t} \myep_j (u_j^{\trans} u_i \myep_i) + \sum_{1\le j < i\le t} (\myep_j u_j^{\trans} u_i) \myep_i + \sum_{1\le i \le t} u_i^{\trans} u_i \myep_i^2  \right] \nonumber\\
&=& \sigma_{\myep}^2\, \E \left[ \sum_{i=1}^t u_i^{\trans} u_i \right] \nonumber\\
&=& \sigma_{\myep}^2\, \E \left[\sum_{i=1}^t  (p_i - p_0)^2 + \sum_{i=1}^t\|\vx_i\|^2 \right] \nonumber\\
&\le & t \sigma_{\myep}^2\,\left[\frac{p_0^2}{2} + \frac{a'^2 + m\,r_{\max}^2 \,x_{\max}^2}{b_{\max}^2}+m \right]\,.\label{eq:upper-1st-term}
\end{eqnarray}
Thus, we see that
\begin{eqnarray}\label{eq:scale}
\E  \left\{[p_t - p^*(\theta,\vx_t) ]^2\right\} &\le&
K_0\,\E\left[\| \theta - \vartheta_t \|^2\right]\nonumber\\
&=&O\left( \frac{1}{t} \right) + O\left([m+1][2e]^{-\frac{\lambda_0 t}{2R}} \right) = O\left( \frac{1}{t} \right)\,.
\end{eqnarray}
Summing up both sides of Eq. \eqref{eq:scale}, we get
\begin{eqnarray}
\Delta_\theta^{\pi^g} (T) &=& -\beta \sum_{t=1}^{T} \E  \left\{[p_t - p^*(\theta,\vx_t) ]^2\right\}
= O\left( \sum_{t=1}^{T} \frac{1}{t} \right) = O\Big(\log(T)\Big)
\end{eqnarray}
in which the summation $\sum_{t=1}^{T} \frac{1}{t}$ is upper bounded by the integral $1 + \int_{1}^{T} \frac{1}{t} dt$. This finishes the proof of $O\Big(\log(T)\Big)$ upper bound on the regret. We can also see that the constant in $O\Big(\log(T)\Big)$ is at most
\begin{equation*}
\frac{4|b_{\min}|\,K_0\,\sigma_{\myep}^2}{\lambda_0^2}\,\left[\frac{p_0^2}{2} + \frac{a'^2 + m\,r_{\max}^2 \,x_{\max}^2}{b_{\max}^2}+m \right]\,,
\end{equation*}
where we used the fact that diagonal entries of $\Sigma_{\vx}$ are $1$ which means its trace is $m$.

Lastly, we should also note that this constant stays finite even when $\gamma=0$. The only dependence on $\gamma$ is through $r_{\max}$ which impacts the regret when $\|\gamma\|_\infty$ becomes large. Thus, we have completed proof of Theorem \ref{thm:greedy} and we only need to prove Lemma \ref{lem:lmin_EuuT}.

\subsubsection{Proof of Lemma \ref{lem:lmin_EuuT}}

Our proof strategy is to show that for any vector $\vy\in\reals^{m+1}$ the following holds
\begin{equation}\label{eq:goal_lmin_proof}
\vy^{\trans}\E[u_iu_i^{\trans}\mid\cF_{i-1}]\,\vy \ge \lambda_0\|\vy\|^2\,.
\end{equation}
To simplify the notation we will drop sub-indices $i$, $i-1$, and $\vx$.
Next we write $\vy=[y_1~\vy_2^{\trans}]^{\trans}$ where $y_1\in\reals$ and $\vy_2\in\reals^m$ which gives
\begin{align}
\vy^{\trans}\E[uu^{\trans}\mid\cF]\,\vy &= \delta^2y_1^2 + (\vq^{\trans}\,\Sigma\,\vq)y_1^2+\vy_2^{\trans}\Sigma\vy_2+2(\vy_2^{\trans}\,\Sigma\,\vq)y_1\nonumber\\
&=\delta^2y_1^2 + (y_1\vq+\vy_2)^{\trans}\,\Sigma \, (y_1\vq+\vy_2)\nonumber\\
&\ge \delta_0^2y_1^2 + \|y_1\vq+\vy_2\|^2\lmin(\Sigma)\,.\label{eq:3}
\end{align}
Combining \eqref{eq:3} with the following inequality
\begin{align*}
\|\vy_2\|^2&\le (\|y_1\vq+\vy_2\| + |y_1|\|\vq\|)^2\le 2\|y_1\vq+\vy_2\|^2 + \frac{r_{\max}^2}{2b_{\max}^2}y_1^2\,,
\end{align*}
we obtain
\begin{align*}
\vy^{\trans}\E[uu^{\trans}\mid\cF\,]\,\vy &\ge
\frac{\delta_0^2}{2}y_1^2 + \min\left[\frac{\delta_0^2b_{\max}^2}{r_{\max}^2},\frac{\lmin(\Sigma)}{2}\right]\left(2\|y_1\vq+\vy_2\|^2 + \frac{r_{\max}^2}{2b_{\max}^2}y_1^2 \right)\\
&\ge \frac{\delta_0^2}{2}y_1^2 + \min\left[\frac{\delta_0^2b_{\max}^2}{r_{\max}^2},\frac{\lmin(\Sigma)}{2}\right]\|\vy_2\|^2\\
&\ge \min\left[\frac{\delta_0^2}{2},\frac{\delta_0^2b_{\max}^2}{ r_{\max}^2},\frac{\lmin(\Sigma)}{2}\right]\|\vy\|^2
\end{align*}
which finishes the proof.

\endproof

\section{Extensions}\label{sec:extensions}

\subsection{Extension to non-iid covariates and shocks}\label{sec:non-iid}

In this section we show that our assumptions on the demand covariates and demand shocks can be generalized.

\paragraph{Demand shocks.} The only assumptions required on the sequence $\{\myep_i\}_{i\ge 1}$ for our proof to go through are that, for all $i\ge1$,
\begin{enumerate}
\item  $\myep_i$ is independent of $\cF_{i-1}$, $p_i$, and $\vx_i$.
\item $\E[\myep_i]=0$.
\item There is a finite constant $\sigma_{\myep}$ such that $\E[\myep_i^2]\le\sigma_{\myep}^2$.
\end{enumerate}
In particular, $\{\myep_i\}_{i\ge 1}$ does not need to be an iid sequence and can be for example a martingale difference sequence with respect to $\{\cF_i\}_{i\ge 0}$ and with finite conditional variance.

\paragraph{Demand covariates.} Similarly, the only assumptions required on the sequence $\{\vx_i\}_{i\ge 1}$ for our proof to go through are that, for all $i\ge1$,
\begin{enumerate}
\item $\E[\vx_i\mid\cF_{i-1}]=0$.
\item $\E[\vx_i\vx_i^{\trans}\mid\cF_{i-1}]=\Sigma_{\vx}^i$ where the sequence of covariance matrices $\{\Sigma_{\vx}^i\}_{i\ge 1}$ are all positive definite with minimum eigenvalues that are uniformly bounded (from below) away from $0$ and maximum eigenvalues that are uniformly bounded from above.
\end{enumerate}

%

\end{APPENDICES}







\bibliographystyle{ormsv080}
\bibliography{mybib}

\end{document}